\documentclass[10pt,journal,compsoc]{IEEEtran}
\ifCLASSOPTIONcompsoc
  
  \usepackage[nocompress]{cite}
\else
 
  \usepackage{cite}
\fi

\usepackage{times}
\usepackage{graphicx}
\usepackage{amsmath}
\usepackage{amssymb}
\usepackage{adjustbox,lipsum}
\usepackage[vlined,boxed,commentsnumbered]{algorithm2e}
\usepackage{subfigure}
\usepackage{algorithmic}
\usepackage{tablefootnote}

\newcommand{\mathbbm}[1]{\text{\usefont{U}{bbm}{m}{n}#1}} 

\ifCLASSINFOpdf

\usepackage{graphicx}
\usepackage{amsmath}
\usepackage{amssymb}
\usepackage{algorithmic}
\usepackage{adjustbox,lipsum}
\usepackage[vlined,boxed,commentsnumbered]{algorithm2e}
\usepackage{subfigure}
\usepackage{tablefootnote}
\usepackage{xcolor}
\usepackage{bm}
\usepackage[font=small,skip=5pt]{caption}
\else

\fi

\newtheorem{definition}{Definition}
\newtheorem{theorem}{Theorem}

\newcommand{\spd}[1]{\mathcal{S}_{++}^{#1}}

\newcommand{\eye}[1]{{I}_{d}}

\newcommand{\reals}[1]{\mathbb{R}^{#1}}

\newcommand{\half}{\frac{1}{2}}

\newcommand{\comment}[1]{}

\newcommand{\ab}{\alpha\beta}

\newcommand{\abldab}[4]{D^{({#3},{#4})}({#1}\parallel{#2})}
\newcommand{\logdet}{\log\det}
\DeclareMathOperator*{\Log}{Log}
\DeclareMathOperator*{\Exp}{Exp}
\newcommand\inv[1]{#1\raisebox{1.15ex}{$\scriptscriptstyle-\!1$}}
\newcommand{\set}[1]{\left\{#1\right\}}
\newcommand{\enorm}[1]{\left\|{#1}\right\|}
\newcommand{\trace}[1]{\mathrm{Tr}\left({#1}\right)}
\newcommand{\grad}[1]{\nabla_{#1}}
\newcommand{\fnorm}[1]{\left\|{#1}\right\|_F}

\DeclareMathOperator*{\diag}{diag}

\newcommand{\bigoh}{\mathcal{O}}
\DeclareMathOperator*{\gridsearch}{Grid Search}

\DeclareMathOperator*{\LEkmeans}{log-euc-kmeans}
\DeclareMathOperator*{\argmin}{arg\,min}
\newcommand{\mX}{X}
\newcommand{\mY}{Y}
\newcommand{\mA}{A}
\newcommand{\vx}{\mathbf{x}}

\newcommand{\sx}{x}
\newcommand{\salpha}{\alpha}
\newcommand{\sbeta}{\beta}
\newcommand{\dataset}{\mathcal{X}}
\newcommand{\labels}{\mathcal{L}}
\newcommand{\dict}{\mathbf{B}}
\newcommand{\mB}{B}
\newcommand{\vv}{\mathbf{v}}
\newcommand{\valpha}{\bm{\alpha}}
\newcommand{\vbeta}{\bm{\beta}}
\newcommand{\mW}{W}
\newcommand{\sy}{y}
\newcommand{\vh}{\mathbf{h}}

\newcommand{\mZ}{Z}

\newcommand{\mH}{H}
\newcommand{\mV}{V}
\newcommand{\mU}{U}
\newcommand{\mC}{C}

\newcommand{\mDelta}{\Delta}
\newcommand{\vdelta}{\bm{\delta}}
\newcommand{\mS}{S}

\newcommand{\tb}[1]{\textbf{#1}}

\newcommand{\BlackBox}{\rule{1.5ex}{1.5ex}}  
\newenvironment{proof}{\par\noindent{\bf Proof\ }}{\hfill\BlackBox\\[2mm]}
\newtheorem{remark}{Remark}

\newcommand{\iddl}{\text{IDDL}}
\newcommand{\idc}{\text{IDC}}

\usepackage{tikz}

\begin{document}
\title{Learning Log-Determinant Divergences for \\Positive Definite Matrices}

\author{Anoop Cherian* \quad Panagiotis Stanitsas* \quad  Jue Wang \quad Mehrtash Harandi \\
\quad Vassilios Morellas \quad Nikolaos Papanikolopoulos
\IEEEcompsocitemizethanks{\IEEEcompsocthanksitem Anoop Cherian* is with Mitsubishi Electric Research Labs (MERL), Cambridge, MA. Work done while at the Australian Centre for Robotic Vision, Australian National University, Canberra. The * indicates equal contribution. E-mail: cherian@merl.com \protect\\
\IEEEcompsocthanksitem Panagiotis Stanitsas* is with the University of Minnesota, Minneapolis, MN. E-mail: stani078@umn.edu\protect\\
\IEEEcompsocthanksitem Jue Wang is with the Research School of Engineering, The
Australian National University, ACT 2601, Australia. E-mail: jue.wang@anu.edu.au\protect\\
\IEEEcompsocthanksitem Mehrtash Harandi is with the department of Electrical  and Computer Systems Engineering, Monash University, and Data61-CSIRO, Melbourne, Australia. E-mail: mehrtash.harandi@monash.edu\protect\\
\IEEEcompsocthanksitem Vassilios Morellas is with the University of Minnesota, Minneapolis, MN. E-mail: morellas@cs.umn.edu\protect\\
\IEEEcompsocthanksitem Nikolaos Papanikolopoulos is with the University of Minnesota, Minneapolis, MN. E-mail: npapas@cs.umn.edu\protect\\

}
\thanks{}}

\markboth{TRANSACTIONS ON PATTERN ANALYSIS AND MACHINE INTELLIGENCE}%
{Shell \MakeLowercase{\textit{et al.}}:Learning $\alpha\beta$-Divergences for Positive Definite Matrices}

\IEEEtitleabstractindextext{%
\begin{abstract}

Representations in the form of Symmetric Positive Definite (SPD) matrices have been popularized in a variety of visual learning applications due to their demonstrated ability to capture rich second-order statistics of visual data. There exist several similarity measures for comparing SPD matrices with documented benefits. However, selecting an appropriate measure for a given problem remains a challenge and in most cases, is the result of a trial-and-error process. In this paper, we propose to learn similarity measures in a data-driven manner. To this end, we capitalize on the $\alpha\beta$-log-det divergence, which is a meta-divergence parametrized by scalars $\alpha$ and $\beta$, subsuming a wide family of popular information divergences on SPD matrices for distinct and discrete values of these parameters. Our key idea is to cast these parameters in a continuum and learn them from data. We systematically extend this idea to learn vector-valued parameters, thereby increasing the expressiveness of the underlying non-linear measure. We conjoin the divergence learning problem with several standard tasks in machine learning, including supervised discriminative dictionary learning and unsupervised SPD matrix clustering. We present Riemannian gradient descent schemes for optimizing our formulations efficiently, and show the usefulness of our method on eight standard computer vision tasks.



\end{abstract}

}

\maketitle
\IEEEdisplaynontitleabstractindextext

\IEEEpeerreviewmaketitle

\section{Introduction}
\label{sec:intro}

The trail of Symmetric Positive Definite (SPD) matrices in computer vision applications is becoming more prominent in the recent years. Examples include, flexible image representations derived in the form of Region CoVariance Descriptors (RCoVDs)~\cite{tuzel2006region} capable of fusing various modalities while compactly capturing their second order statistics. Popular deep learning architectures have also benefited by the use of SPD matrices employing them as second-order pooling operators~\cite{cherian2019second,gao2016compact,ionescu2015matrix,huang2017RiemAAAI}. In an effort to harness the representational power of SPD matrices, the newly devised domain of geometric deep learning has produced powerful machinery for addressing inference problems in a Riemannian deep learning setup~\cite{huang2017RiemAAAI}. More classical approaches promote the use of SPD matrices as kernel matrices of high-dimensional data~\cite{harandi2015riemannian}, points in diffusion MRI~\cite{pennec2006}, and diffusion tensors~\cite{brox2006nonlinear}. 

SPD matrices belong to an open cone in the Euclidean space as a result of their positive definiteness property, and thus a natural way to compare two SPD matrices is the standard Euclidean distance. However, it is often found practically that using a Euclidean geometry leads to sub-optimal performance. A motivating application in this regard is perhaps the problem of computing the barycenter of a set of SPD matrices~\cite{yger2015averaging,bini2013computing}; a common task arising in ensemble learning of Gaussian models~\cite{lakshminarayanan2017simple} or when interpolating DT-MRI points (which are SPD matrices)~\cite{arsigny2006log,pennec2006}. It is often found that using a Euclidean geometry in these applications lead to unrealistic barycenters (often producing bloated SPD matrices, which may be interpreted as an overestimated covariance in Gaussian models). However, enforcing a non-linear (often Riemannian) geometry on the SPD cone using a suitable non-linear measure avoids such pitfalls, leading to practical benefits, as demonstrated in several successful applications~\cite{ harandi2014manifold,cherian2019second, wang2012covariance,gao2016compact,yu2018statistically,lin2018second}.

Interestingly, SPD matrices constitute a very rich class of mathematical objects that appear in several disciplines, and for which a wide variety of similarity measures and geometries have been associated with. Notable representatives of such measures include: (i) the Affine Invariant Riemannian Metric (AIRM) which is a geodesic distance induced by their natural Riemannian geometry~\cite{pennec2006}, (ii) the Jensen-Bregman~\cite{cherian2013jensen,sra2012new,bhattacharyya1943measure} and Burg~\cite{kulis2006learning} divergences which result from information geometry perspective and, (iii) the Jeffreys KL divergence (KLDM) using relative entropy~\cite{moakher2006symmetric}, among several others such as the Bures distance used in quantum information theory~\cite{luo2004informational}, Bures-Wasserstein distance~\cite{bhatia2019bures} recently proposed in optimal transport theory, the popular log-Euclidean Riemannian metric used in diffusion MRI~\cite{arsigny2006log} and its kernel analogues such as the Hilbert-Schmidt metric~\cite{quang2014log}. Given such an elaborate choice of potential measures, it is overwhelming to consider choosing an appropriate similarity for a given problem.  


In this paper, we attempt to tackle the challenge of choosing the similarity measure using data-driven approach by directly learning the divergence from a training set of SPD matrices. While, one may resort to standard metric learning~\cite{davis2007information} ideas or their SPD matrix variants~\cite{huang2017riemannian,huang2015log,zadeh2016geometric,harandi2014manifold} in this regard, we propose to use the recently introduced $\alpha\beta$-Logdet Divergence (ABLD)~\cite{cichocki2015log} for this purpose. ABLD is a meta-divergence on SPD matrices, parametrized by scalars $\alpha$ and $\beta$, and which is shown to be exactly equal to some of the popular divergences and metrics listed above for specific values of these scalars. For example, ABLD converges to AIRM as $\alpha,\beta\rightarrow 0$ and is equal to Burg divergence when $\alpha=\beta=1$, among others (see Sec.~\ref{sec:background}). This unifying result suggests that we could learn to choose an appropriate similarity measure in a data-driven way by learning $\alpha$ and $\beta$. In contrast to standard metric learning approaches, our proposed learning setup not only allows learning a suitable measure from a continuum of log-det divergences, but also guarantees that standard measures are included in the learning landscape.

While, learning ABLD is our primary motivation, we go far beyond this goal and generalizes our learning setup in two predominant ways, (i) a dictionary learning and divergence learning setting for supervised regression and classification, and (ii) divergence learning combined with unsupervised clustering in a K-Means setting. Specifically, instead of learning the parameters of a single ABLD, we propose to learn a family of ABLDs, each parametrized separately, via learning a vector of parameter pairs under the assumption that there might be distinct SPD matrix subspaces (clusters or classes) in the given data that may adhere to their own divergences. We further parameterize the origin in these subspaces via learning them; our full setup we call \emph{Information Divergence and Dictionary Learning} (IDDL), where each origin forms an atom in an SPD matrix dictionary. Using this setup, we propose SPD data regression and classification models by embedding a given SPD matrix into a vector; each dimension of this vector capturing its similarity to a respective dictionary atom via its learned divergence. We combine IDDL with i) a ridge-regression objective, and ii) using a structured-SVM objective. Our full model (including IDDL and classifier) are learned end-to-end in a Riemannian alternating minimization setup. For clustering, we derive a scheme for K-Means on SPD matrices while addressing the problem of finding optimal values of $\alpha$ and $\beta$. This effort yields a variant of K-Means, which we call $\alpha\beta$-KMeans. 



To evaluate our frameworks, we present experiments on an extensive set of computer vision applications, including texture recognition, activity recognition, 3D object, and cancerous tissue recognition, on a multitude of datasets with different levels of data complexities. We supplement our experimental comparisons with an equally extensive ablation study of our schemes under various settings. Our results demonstrate the benefits of joint learning by achieving state-of-the-art performance against competing techniques, including the recent sparse coding, Riemannian metric learning, and kernel coding schemes. 

This paper extends our previous works on this topic~\cite{stanitsas2017clustering, anoop_panos2017}, which consider the IDDL and AB-KMeans clustering setups separately. In this paper, we not only unifies these setups, but also explores more richer classification settings via a structured-SVM formulation, which is jointly learned with our IDDL framework. We also provide additional technical details of our optimization. Extensive discussion is provided in the form of a detailed ablation study discussing all the different elements of our scheme.

\section{Related Work}
\label{sec:related_work}
The $\ab$-logdet divergence is a matrix generalization of the well-known $\ab$-divergence~\cite{cichocki2010families} that computes the (a)symmetric (dis)similarity between two finite positive measures (data densities). As the name implies, $\ab$-divergence is a unification of the so-called $\alpha$-family of divergences~\cite{amari2007methods} (that includes popular measures such as the KL-divergence, Jensen-Shannon divergence, and the chi-square divergence) and the $\beta$-family~\cite{basu1998robust} (including the squared Euclidean distance and the Itakura Saito distance). Against several standard measures for computing similarities, both $\alpha$ and $\beta$ divergences are known to lead to solutions that are robust to outliers and additive noise~\cite{lafferty1999additive}, thereby improving application performance. They have been used in several statistical learning applications including non-negative matrix factorization
~\cite{cichocki2009nonnegative,kompass2007generalized,dhillon2005generalized}, nearest neighbor embedding~\cite{hinton2002stochastic}, and blind-source separation~\cite{mihoko2002robust}.  

A class of methods with similarities to our formulation are metric learning schemes on SPD matrices. One popular technique is the manifold-manifold embedding of large SPD matrices into a lower-dimensional SPD space in a discriminative setting~\cite{harandi2014manifold}. Log-Euclidean metric learning has also been proposed for this embedding in~\cite{huang2015log,sivalingam2009metric}. While, we also learn a metric in a discriminative setup, ours is different in that we learn an information divergence. In Thiyam et al.~\cite{thiyam2017optimization}, ABLD is proposed replacing symmetric KL divergence in better characterizing the learning of a decision hyperplane for BCI applications. In contrast, we propose to embed the data matrices as vectors, each dimension of these vectors learning a different ABLD, thus leading to a richer representation of the input matrix.  More recently, extensions of ABLD to an infinite dimensional Hilbert space setting is explored in~\cite{minh2019alpha}. However, our work is complementary to this effort, and explores a finite-dimensional setting.

Vectorial embedding of SPD matrices has been investigated using disparate formulations for computer vision applications. As alluded to earlier, the log-Euclidean projection~\cite{arsigny2006log} is a common way to achieve this, where an SPD matrix is isomorphically mapped to the Euclidean space of symmetric matrices using the matrix logarithm. Popular sparse coding schemes have been extended to SPD matrices in~\cite{cherian2016riemannian, sivalingam2010tensor,xie2013nonlinear}
 using SPD dictionaries, where the resulting sparse vector is assumed Euclidean. Another popular way to handle the non-linear geometry of SPD matrices is to resort to kernel schemes by embedding the matrices in an infinite dimensional Hilbert space which is assumed to be linear ~\cite{harandi2014bregman,li2013log,harandi2015riemannian}. In all these methods, the underlying similarity measure is fixed and is usually chosen to be one among the popular $\ab$-logdet divergences or the log-Euclidean metric. 
 
 \begin{table*}[htbp]
\centering
\begin{tabular}{|c|c|c|}
\hline
$(\alpha,\beta)$  & ABLD & Divergence\\
\hline
$(\alpha,\beta)\rightarrow 0$ & $\fnorm{\Log{\mX^{-\half}\mY\mX^{-\half}}}^2$ &  Squared Affine Invariant Riemannian Metric (AIRM)~\cite{pennec2006} \\
\hline
$\alpha=\beta=\pm\half$ &  $4\left(\logdet\frac{\mX+\mY}{2}-\half\logdet\mX\mY\right)$ & Jensen-Bregman Logdet Divergence (JBLD)~\cite{cherian2013jensen} \\
\hline
$\alpha=\pm 1,\beta\rightarrow 0$ & $\half\trace{\mX\inv{\mY} + \mY\inv{\mX}} - d$ & Jeffreys KL Divergence\footnote{using the symmetrization of ABLD.}~\cite{moakher2006symmetric}\\
\hline
$\alpha=1,\beta=1$ & $\trace{\mX\inv{\mY}} - \logdet\mX\inv{\mY} - d$ & Burg Matrix Divergence~\cite{kulis2006learning} \\
\hline
\end{tabular}
\caption{ABLD and its connections to popular divergences.}
\label{tab:1}
\end{table*}
 
Several unsupervised schemes for clustering SPD matrices have also been proposed in the relevant literature. Commonly used schemes capitalize on conventional clustering machinery after being modified towards abiding to the non-linear geometry of SPD matrices. In this direction, two extensions of the popular KMeans have been derived admitting the manifold of the SPD matrices. In the first variant of KMeans, centroids are computed using the Karcher means algorithm~\cite{bini2013computing} and the affine-invariant Riemannian metric~\cite{pennec2006}. Substituting the similarity computation based on the AIRM by the log-Euclidean metric~\cite{arsigny2006log}, yields a second variant of KMeans for SPD matrices termed LE-KMeans. Using the matrix logarithm operation, which entails a diffeomorphic mapping of an SPD matrix onto its tangent space, allows for distance computations in a Euclidean manner (as this tangent space is  Euclidean). In that way, centroids are computed by averaging the samples' vectorial representations in the tangent space. Additional variants of KMeans can be derived by capitalizing on the different similarity measures for SPD matrices. 

A second family of clustering schemes for SPD matrices takes advantage of Euclidean embeddings in the form of similarity matrices computed using suitable measures. In that direction, Spectral clustering schemes have been developed for SPD matrices by computing suitable Mercer kernels on the data using appropriate distances (e.g., LE). Sparse subspace clustering schemes have also been derived for SPD matrices via their embedding into a Reproducing Kernel Hilbert Space~\cite{yin2016kernel, patel2014kernel}. Such schemes come at the expense of additional memory requirements involved with computing the eigen spectrum of the computed kernel. 

In addition, non-parametric schemes for clustering SPD matrices have been derived in the form of dimensionality reduction on Riemannian manifolds~\cite{goh2008clustering}, using Locally Linear Embeddings~\cite{roweis2000nonlinear}, or capitalizing on the Laplacian eigenmaps~\cite{belkin2002laplacian}. In the family of non-parametric clustering algorithms, a Bayesian framework for SPD matrices is formulated using the Dirichlet Process~\cite{cherian2016bayesian}. Finally, variants of the Mean shift clustering algorithm and Kernel Density Estimation for SPD matrices have also been derived in~\cite{subbarao2009nonlinear} and~\cite{pelletier2005kernel} respectively. 

In contrast to all these methods, to the best of our knowledge, this is the first unified attempt to bridge the learning of information divergences with dictionary learning and clustering objectives. Even though automatic selection of the parameters of $\ab$-divergence is investigated in~\cite{csimcsekli2015learning,dikmen2015learning} they deal solely with scalar density functions in a maximum-likelihood setup and do not consider the optimization of $\alpha$ and $\beta$ jointly. 

\textbf{Notation:} Following standard practice, we use upper case for matrices (such as $\mX$), lower-bold case for vectors $\vx$, and lower case for scalars $\sx$. Further, $\spd{d}$ is used to denote the cone of $d\times d$ SPD matrices. We use $\dict$ to denote a 3D tensor each slice of which is an SPD matrix of size $d\times d$. Further, we use $\eye{d}$ to denote the $d\times d$ identity matrix, $\Log$ for the matrix logarithm, and $\diag$ for the diagonalization operator. In addition, we use $\mathbf{C}$ to denote a 3D tensor each slice of which corresponds to an SPD centroid of size $d\times d$ for the proposed clustering setup. Finally, we use $\mathbf{\Pi}= \{\pi_1, ..., \pi_k\}$ to denote a clustering of data into $k$ partitions; $\pi_i$ is the $i$-th partition and comprises a subset of the dataset assigned to this cluster.

\section{Background}
\label{sec:background}
In this section, we setup the mathematical preliminaries necessary to elucidate our contributions. 

\subsection{Alpha-Beta-Log-Determinant Divergence (ABLD)}
Introduced in Cichocki et al.~\cite{cichocki2015log}, the $\alpha\beta$-log-det divergence is a result of an effort to extend the log-det divergences, existing in various forms~\cite{pennec2006,cherian2013jensen,moakher2006symmetric}, and finding connections between them. Below, we review this meta-divergence, its connections to other divergences, and some of its properties that will be useful in our formulations. 
\begin{definition}[ABLD~\cite{cichocki2015log}]
For $X,Y\in\spd{d}$, the $\ab$-log-det divergence is defined as:
\begin{equation}
	D^{(\alpha,\beta)}\hspace*{-0.3ex}(\mX \hspace*{-0.5ex} \parallel \hspace*{-0.5ex} \mY)
	\hspace*{-0.5ex} = \hspace*{-0.5ex} \frac{1}{\salpha\sbeta}\logdet \hspace*{-0.5ex}
	\left(\frac{\salpha(\mX\mY^{-1})^{\sbeta} \hspace*{-0.5ex} + \hspace*{-0.5ex}\sbeta(\mX\mY^{-1})^{-\salpha}}
	{\salpha+\sbeta}\right)\hspace*{-0.5ex},
	\label{eq:abld}	
\end{equation}
\begin{equation}
     \salpha\neq 0,\ \sbeta\neq 0\ \text{ and } \salpha+\sbeta\neq 0.
     \label{eq:abld_constraints}
\end{equation}
\end{definition}

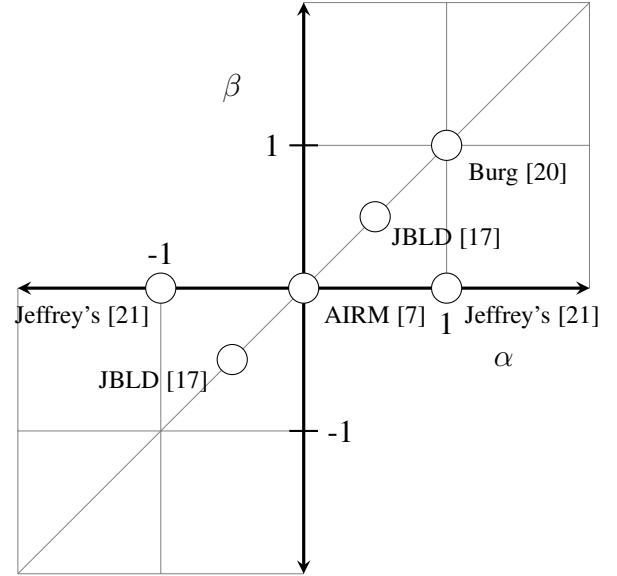
\begin{figure}[tbp]
\begin{center}
\begin{tikzpicture}[scale=1.9]
  \draw[very thin, color=gray] (0,0) grid (2,2);
  \draw[very thin, color=gray] (0,0) grid (-2,-2);
  \draw[very thin, color=gray] (0,0) -- (-2,-2);
  \draw[very thin, color=gray] (0,0) -- (2,2);
  \draw [very thick, -stealth] (0,0) -- (2,0);
  \draw [very thick, -stealth] (0,0) -- (0,2);
  \draw [very thick, -stealth] (0,0) -- (-2,0);
  \draw [very thick, -stealth] (0,0) -- (0,-2);
  \node (a) at (1.4,-0.5) {\large $\alpha$};
  \node (a) at (-0.5,1.4) {\large $\beta$};
  \draw [thick] (1,0.1) -- (1,-0.1) node [below] {\large 1};
  \draw [thick] (-1,-0.1) -- (-1,0.1) node [above] {\large -1};
  \draw [thick] (0.1,1) -- (-0.1,1) node [left] {\large 1};
  \draw [thick] (-0.1,-1) -- (0.1,-1) node [right] {\large -1};
  \draw [fill=white] (1,1) circle (3pt);
  \node (b) at (1.5,0.8) {Burg~\cite{kulis2006learning}};
  \draw [fill=white] (0,0) circle (3pt);
  \node (b) at (0.5,-0.2) {AIRM~\cite{pennec2006}};
  \draw [fill=white] (1,0) circle (3pt);
  \node (b) at (1.6,-0.2) {Jeffrey's~\cite{moakher2006symmetric}};
  \draw [fill=white] (0.5,0.5) circle (3pt);
  \node (b) at (1,0.35) {JBLD~\cite{cherian2013jensen}};
  \draw [fill=white] (-0.5, -0.5) circle (3pt);
  \node (b) at (-1.05,-0.65) {JBLD~\cite{cherian2013jensen}};
  \draw [fill=white] (-1,0) circle (3pt);
  \node (b) at (-1.55,-0.2) {Jeffrey's~\cite{moakher2006symmetric}};
\end{tikzpicture}
\caption{An illustration of ABLD and its connections to popular divergences on SPD matrices.\label{fig:abld_connections}}
\end{center}

\end{figure}

It can be shown that ABLD depends only on the generalized eigenvalues of $\mX$ and $\mY$~\cite{cichocki2015log}. 
Suppose $\lambda_i$ denotes the $i$-th eigenvalue of $\mX\inv{\mY}$. Then under constraints defined in~\eqref{eq:abld_constraints}, 
we can rewrite~\eqref{eq:abld} as:
\begin{equation}
	D^{(\alpha,\beta)}\hspace*{-0.3ex}(\mX \hspace*{-0.7ex} \parallel \hspace*{-0.5ex} \mY \hspace*{-0.25ex})
    \hspace*{-0.5ex} = \hspace*{-0.5ex} \frac{1}{\salpha\sbeta} \hspace*{-0.5ex} \sum_{i=1}^d 
    \hspace*{-0.25ex} \log \Big(\salpha\lambda_i^{\sbeta} 
    \hspace*{-0.5ex} + \hspace*{-0.3ex} \sbeta \lambda_i^{-\salpha} 
    \Big)
    \hspace*{-0.25ex} - \hspace*{-0.25ex} d\log\left(\salpha \hspace*{-0.5ex} + \hspace*{-0.5ex} \sbeta \right)\hspace*{-0.5ex}.
    \label{eq:abld_lambda}
\end{equation}
 This formulation will come handy when deriving the gradient updates for $\alpha$ and $\beta$ in the sequel. As alluded to earlier, a hallmark of ABLD is that it unifies several popular distance measures on SPD matrices that one commonly encounters in machine learning and vision applications. In Table~\ref{tab:1}, we provide connections of ABLD with popular similarity measures on SPD matrices. A graphical illustration of the landscape of these measures is provided in Figure~\ref{fig:abld_connections} for various values of $\alpha$ and $\beta$.

\subsection{ABLD Properties}
\noindent\textbf{Avoiding Degeneracy:} An important observation regarding the design of optimization algorithms on ABLD is that the quantity inside the $\logdet$ term has to be positive definite. When both $\alpha,\beta$ are simultaneously positive or negative, this quantity is positive-definite. However, when $\alpha$ and $\beta$ have different signs, the $\logdet$ term may become degenerate, the necessary conditions to avoid which are stipulated by the following theorem.
\begin{theorem}
For $\mX,\mY\in\spd{d}$, if $\lambda_i$ is the $i$-th eigenvalue of $\mX\inv{\mY}$, then $\abldab{\mX}{\mY}{\salpha}{\sbeta}\geq 0$ only if 
\begin{align}
\lambda_i & > \left|\frac{\alpha}{\beta}\right|^{\frac{1}{\alpha+\beta}}\!\!\!, \text{ for } \alpha>0 \text{ and } \beta < 0, \text{ or } \\
\lambda_i &< \left|\frac{\beta}{\alpha}\right|^{\frac{1}{\alpha+\beta}}\!\!\!,\text{ for } \alpha<0 \text{ and } \beta > 0, \forall i=1,2,\cdots, d.
\end{align}
\begin{proof}
See~\cite{cichocki2015log}.
\end{proof}
\label{thm:1}
\end{theorem}
As the algorithms we introduce in the sequel use sets of input SPD matrices, it may be computationally challenging to impose the restrictions in Theorem~\ref{thm:1} on the learned $\alpha$ and $\beta$ for every matrix pair. Thus, we restrict our scope in this paper to the form of ABLD when both $\alpha$ and $\beta$ have the same sign, thereby avoiding the quantity inside $\logdet{}$ to become indefinite. We make this assumption in our formulations in Section~\ref{sec:proposed_method}.

\noindent\textbf{Affine Invariance:} For a non-singular matrix $\mathbf{A} \in \mathbb{R}^{d \times d}$, ABLD is invariant under congruent transformations:
\begin{equation}
\abldab{\mX}{\mY}{\salpha}{\sbeta} = \abldab{\mA\mX\mA^\top}{\mA\mY\mA^\top}{\salpha}{\sbeta}.
\end{equation}
\noindent This is an important property that makes this divergence useful in a variety of applications, e.g., diffusion MRI~\cite{wang2004affine}, where affine-invariance allows the similarity to be robust to anatomical changes in the subjects or to configurations of the image acquisition hardware. Assuming affine invariance also helps derive a metric in a Riemmanian symmetric space; this metric is often found to be better in restoration, interpolation, and filtering of DTMRI images~\cite{pennec2006}.

\noindent\textbf{Identity of Indiscernibles:} For any $\alpha,\beta$, it holds that
\begin{equation}
    \abldab{\mX}{\mY}{\salpha}{\sbeta} = 0 \ \ \textrm{if and only if} \ \ \mX = \mY.
\end{equation}

\noindent\textbf{Scaling Invariance:} For any $c > 0$, it is easy to show that
\begin{equation}
    \abldab{c \mX}{c \mY}{\salpha}{\sbeta} = \abldab{\mX}{\mY}{\salpha}{\sbeta}.
\end{equation}

\noindent\textbf{Smoothness of $\alpha$, $\beta$:}
Assuming $\alpha,\beta$ have the same sign, except at the origin ($\alpha=\beta=0$), ABLD is smooth everywhere with respect to $\alpha$ and $\beta$, thus allowing us to develop Newton-type algorithms on them. Due to the discontinuity at the origin, we ought to design algorithms specifically addressing this particular case.

\noindent\textbf{Dual Symmetry:}
This property allows us to extend results derived for the case of $\alpha$  to the one on $\beta$ later.
\begin{equation}
\abldab{\mX}{\mY}{\salpha}{\sbeta} = \abldab{\mY}{\mX}{\sbeta}{\salpha}.
\label{eq:dual_sym}
\end{equation}

\section{Proposed Method}
\label{sec:proposed_method}
We start by introducing our \emph{Information Divergence and Dictionary Learning} (IDDL) problem setup in which a discriminative divergence learning framework is combined with a dictionary learning setting. Specifically, instead of using a single divergence measure on the SPD matrices, we present a very general setting in which we assume there are SPD subspaces in the data, which may be characterized via learning separate ABLD measures, each parametrized distinctly. We further generalize this model into a bag-of-words setting, by combining our multi-divergence learning problem with an SPD dictionary learning problem, where we also assume that each divergence also has its own SPD subspace origin with respect to which the divergence is measured. We include our IDDL framework within a joint classification setup. We explore two variants of the classification model: i) using a ridge-regression loss, and ii) using a structured SVM loss. Next, we present divergence learning within a K-Means clustering setup, where alongside learning the ABLD parameters, we also learn the SPD cluster centroids. 



\begin{figure}[t]
\centering
\includegraphics[width=8cm,trim={8.9cm 0cm 2cm 0cm},clip]{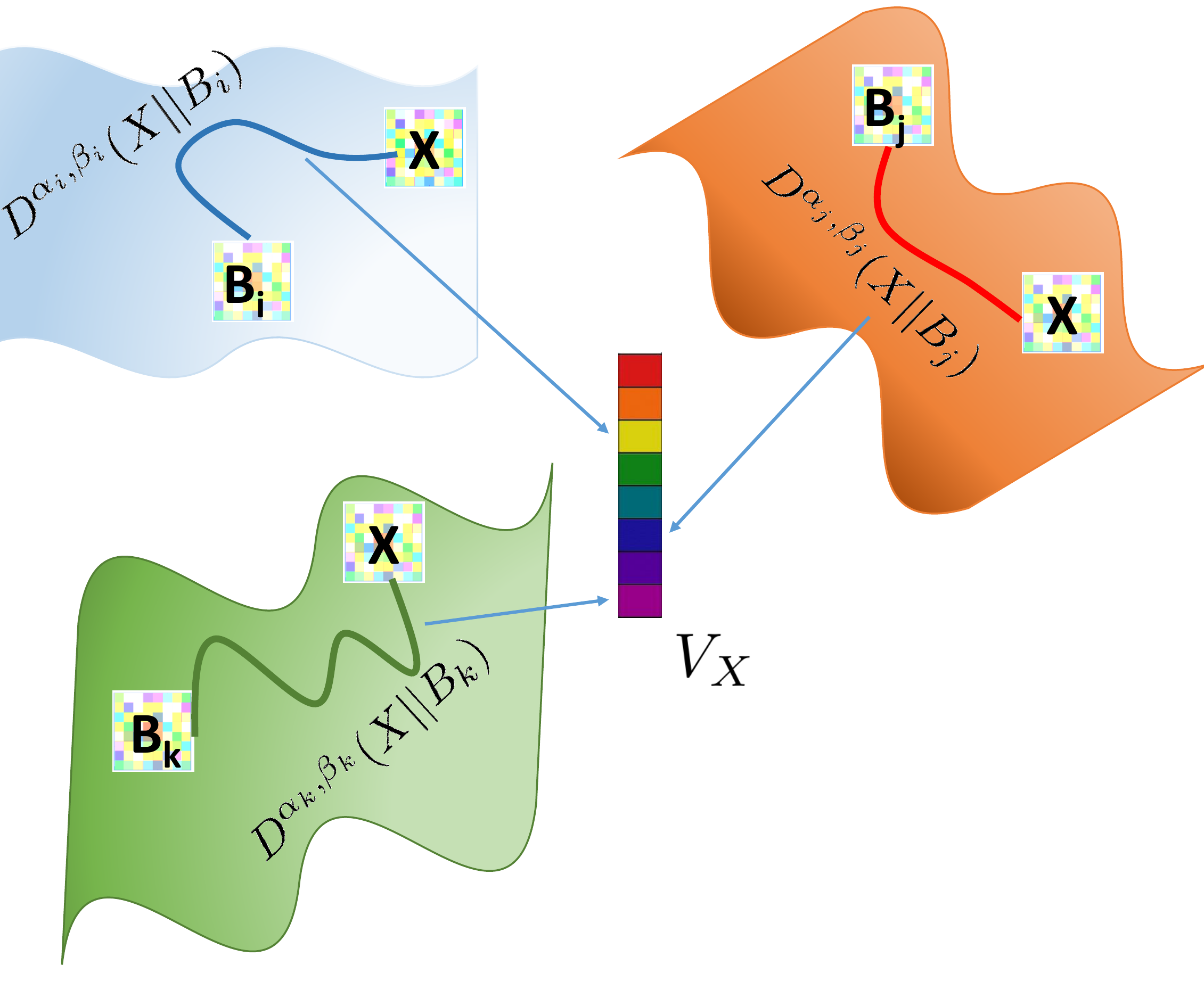}
\caption{A schematic illustration of our IDDL scheme. From an infinite set of potential geometries, our goal is to learn multiple geometries (parameterized by $(\alpha,\beta)$) and representative dictionary atoms for each geometry (represented by $B$'s), such that a given SPD data matrix $X$ can be embedded into a similarity vector $V_X$, each dimension of which captures the divergence of $X$ to the $B$s using the respective measure. We use $V_X$ for classification.}
\label{fig:iddl}
\end{figure}

\subsection{Information Divergence \& Dictionary Learning}

Suppose we are given a dataset of SPD matrices $\dataset=\set{\mX_1,\mX_2,\cdots, \mX_N}$, each $~\mX_i\in\spd{d}$ and their associated class labels $\sy_i\in\labels=\set{1,2,\cdots, L}$. Our goal is to amalgamate three learning pursuits, namely,  (i) learn a dictionary $\dict\in\spd{d}\times_n$, a product of $n$ SPD manifolds, (ii) learn an ABLD on each dictionary atom to best represent the given data for the task of classification, and (iii) learn a discriminative objective function on the encoded SPD matrices (in terms of $\dict$ and the respective ABLDs) for the purpose of classification. This is the most general form of our classification model. For example, if we assume the dictionary atoms are all identity matrices, then our setting is basically a multi-divergence learning setting, while if we assume $n=1$, we have a single divergence learning setup. We formalize the three aforementioned objectives in the formulation presented below.
\begin{align}
\label{eq:iddl}
\iddl := \min_{\substack{\dict>0,\mW,\\\valpha>0,\vbeta>0}}\quad &  \sum_{i=1}^N  f(\vv_i, \sy_i; \mW) \notag\\
\text{subject to } \vv^k_i &= \abldab{\mX_i}{\mB_k}{\valpha_k}{\vbeta_k},
\end{align}

\noindent where the $k$-th dictionary atom in $\dict$ is denoted by $\mB_k$, the vector $\vv_i\in\reals{n}$ denotes the encoding of $\mX_i$ in terms of the dictionary, and $\vv_i^k$ is the $k$-th dimension of this encoding. Specifically, the $k$-th dimension of $\vv_i$ captures the distance of $\mX_i$ to the dictionary atom $\mB_k$ via the respective divergence $\abldab{\mX_i}{\mB_k}{\valpha_k}{\vbeta_k}$. Note that, in our formulations, we assume $\alpha,\beta$ are non-negative to avoid degenerate solutions during optimization caused by the constraint in~\eqref{eq:abld_constraints}. Instead, we consider the positive and negative orthants separately, and design specialized descent for the case when $\alpha=\beta=0$, and select the parameter configuration that leads to the smallest loss. The function $f$ in~\eqref{eq:iddl}, parametrized by $\mW\in\reals{L\times n}$, learns a classifier on $\vv_i$ according to the provided class labels $\sy_i$. Figure~\ref{fig:iddl} illustrates our classification model. For the classifier, we consider two popular loss functions, (i) a ridge regression loss, which has a closed form update as we show below, and (ii) a structured-SVM loss which we optimize in an iterative manner.

\subsubsection{\textbf{Ridge Regression Loss}}
Our first choice for $f$ is a simple ridge regression objective. Let $\vh_i\in\{0,1\}^L$ be a one-off encoding of class labels (i.e., $\vh_i^{\sy_i} = 1$, zero everywhere else). Then, we define $f_1$ as follows, 

\begin{equation}
f_1(\vv_i, \sy_i; \mW) = \frac{1}{2}\enorm{\vh_i - \mW\vv_i}^2 + \gamma\fnorm{\mW}^2,
\label{eq:ridge}
\end{equation}
\noindent where $\gamma$ is a regularization parameter. 

\subsubsection{\textbf{Structured-SVM Loss}}
A more richer choice for $f$ in~\eqref{eq:iddl} is perhaps to consider a max-margin hinge loss (similar to structured-SVM). Letting $\Delta$ denote a control variable on the margins of the classifiers and let $g(\mathbf{v}_i, W) = W\mathbf{v}_i + b$, for a bias $b\in\reals{L\times 1}$, we define $f_2$ as:

\begin{align}
    f_2(\mathbf{v}_i, y_i; W) &= \sum_{l \neq y_i} \left( \max\left( 0, g(\mathbf{v}_i,W )_l - g(\mathbf{v}_i, W)_{y_i} + \Delta\right) \right) \nonumber \\ &\qquad\qquad+ \gamma ||W||_F^2,
    \label{eq:svm}
\end{align}
\noindent where $\gamma$ is a regularization parameter, and $g(\mathbf{v}_i, W)_{y_i}$ represent the $y_i$ element of the vector $g$. For simplifying our notation, in the sequel, for the case of structured-SVM loss, we assume $\mW\in\reals{L+1\times n}$, where the last row of $\mW$ captures the bias $b$. Similarly, we augment $\vv$ to have $n+1$ dimensions, where the last dimension is a constant with value one.

\subsection{Information Divergence \& Clustering}
In the clustering setup, our objective is to partition a set of SPD matrices $\dataset=\set{\mathbf{X}_1,  \mathbf{X}_2, \cdots, \mathbf{X}_N}$ into $k$ partitions, for a given $k$. Let $\mathbf{\Pi} = \set{\pi_1, \cdots, \pi_k}$ denote a partitioning of $\dataset$ where $\pi_i$ is the set of samples assigned to the $i$-th cluster and let $\mathbf{C}_i$ be the respective cluster centroid. We cast the joint \emph{information divergence learning and clustering} (IDC) problem as:
\begin{align}
\label{eq:abkmeans}
 \textrm{IDC}:= \min_{\mathbf{C}, \mathbf{ \Pi}, \alpha, \beta>0} &  f_3\left(\mathbf{\Pi}, \mathcal{X}; \alpha, \beta\right) + \mathbf{\Omega}(\alpha, \beta),
\end{align}
where our objective jointly learns a partition of data points $\mathbf{\Pi}$, the cluster centroids $\mathbf{C}$, and the divergence scalar parameters $\alpha,\beta$. The function $\mathbf{\Omega}(\alpha, \beta) = \mu \left(\alpha^2 + \beta^2\right)$ is a regularization term on the parameters $\alpha$ and $\beta$, and $\mu$ is a regularization constant. 
Substituting the standard KMeans formulation in~\eqref{eq:abkmeans} and using ABLD as the similarity measure, we have the following definition for $f_3$:
\begin{align}
f_3\left(\mathbf{\Pi}, \mathcal{X}; \alpha, \beta\right)=\sum_{\pi \in \mathbf{\Pi}}\sum_{ i \in \pi}  \left(\abldab{\mX_i}{\mathbf{C}_{\pi}}{\alpha}{\beta} \right).
\label{eq:ab_clustering}
\end{align}

\section{Efficient Optimization}
\label{sec:efficient_optimization}
In this section, we briefly review the necessary Riemannian optimization machinery that we resort to for solving our objectives formulated in the last section.

\subsection{Optimization on Riemannian Manifolds}
\label{sec:rcg}

As was shown in Section~\ref{sec:proposed_method}, we need to solve a non-convex constrained optimization problem in the form:
\begin{align}
\mathrm{minimize}~\mathcal{L}(B) \notag\\
\mathrm{s.t.}~~~B \in \spd{d}\;.
\label{eqn:opt_riemannian_manifold}
\end{align}

Classical optimization methods generally turn a constrained problem into a sequence of unconstrained
problems for which unconstrained techniques can be applied.
In contrast, in this paper we make use of the optimization on Riemannian manifolds to 
minimize~\eqref{eqn:opt_riemannian_manifold}. 
This is motivated by recent advances in Riemannian optimization techniques where benefits of exploiting geometry over standard constrained optimization are shown~\cite{absil2009optimization}. As a consequence, these techniques have become increasingly popular in diverse application domains~\cite{cherian2016riemannian,harandi2015riemannian}. 

A detailed discussion of Riemannian optimization goes beyond the scope of this paper, and we refer the interested reader 
to~\cite{absil2009optimization}. 
However, the knowledge of some basic concepts will be useful in the remainder of this paper. As such, here, we briefly consider the case of Riemannian Conjugate Gradient method (RCG), our choice when the empirical study of this work is considered. First, we formally define the SPD manifold.

\begin{definition}[The SPD Manifold]
The set of ($d \times d$) dimensional real, SPD matrices endowed with the Affine Invariant Riemannian Metric (AIRM)~\cite{pennec2006} forms the  SPD manifold $\spd{d}$. 
\begin{equation}
\spd{p} \triangleq \{{X} \in \mathbb{R}^{d \times d}: {v}^T {X} {v} >0,~\forall {v} \in \mathbb{R}^d-\{{0}_d\}\}\;.
\label{eqn:spd_manifold}
\end{equation}
\end{definition}

To minimize~\eqref{eqn:opt_riemannian_manifold}, RCG starts from an initial solution $\mB^{(0)}$ and improves 
its solution using the update rule
\begin{equation}
\mB^{(t+1)} = \tau_{\mB^{(t)}} 
\big( P^{(t)}
\big)\;,
\label{eqn:rcg_retraction}
\end{equation}
where  $P^{(t)}$ identifies a search direction and $\tau_{\mB} ( \cdot): T_\mB\spd{d} \to \spd{d}$
is a \emph{retraction}. The retraction serves to identify the new solution along the geodesic defined by the search direction $P^{(t)}$.
In RCG, it is guaranteed that the new solution obtained by Eq.~\eqref{eqn:rcg_retraction} is on $\spd{d}$ and has a lower objective. The search direction $P^{(t)} \in T_{\mB^{(t)}}\spd{d}$ is obtained by
\begin{equation}
P^{(t)} = -\mathrm{grad}~\mathcal{L}(\mB^{(t)}) + \eta^{(t)} \pi(P^{(t-1)},{\mB^{(t-1)} , \mB^{(t)}})\;.
\label{eqn:rcg_search_direction}
\end{equation}

Here, $\eta^{(t)}$ can be thought of as a variable learning rate, obtained via techniques such as 
Fletcher-Reeves~\cite{absil2009optimization}. Furthermore, $\mathrm{grad}~\mathcal{L}(\mB)$ is the Riemannian gradient of the objective function
at $\mB$ and $\pi(P,\mX , \mY)$ denotes the parallel transport of $P$ from $T_\mX$ to $T_\mY$. In Table~\ref{tab:riemannian_tools}, we define the mathematical entities required to perform RCG on the SPD manifold. Note that computing the standard Euclidean gradient of the function $\mathcal{L}$, denoted by $\nabla_{*}(\mathcal{L})$, is the only requirement to perform RCG on $\spd{d}$.
%

\begin{table}[ht]
\small
\centering
\begin{tabular}{|c|l|}
\hline
{\bf } &{\bf \qquad\qquad\qquad $\spd{d}$}\\
\hline 
\textbf{Riemannian gradient}				&$\mathrm{grad}~\mathcal{L}(\mB) = \mB \mathrm{sym}\big(\nabla_B(\mathcal{L})\big) \mB$\\
\textbf{Retraction.}						&$\tau_\mB(\xi) = \mB^\frac{1}{2}\Exp(\mB^{-\frac{1}{2}}\xi \mB^{-\frac{1}{2}})\mB^\frac{1}{2}$\\
\textbf{Parallel Transport.}				&$\pi(P,\mX , \mY) = Z P Z^T$\\
\hline
\end{tabular}
\vspace{1ex}
\caption{Riemannian tools to perform RCG on $\spd{d}$. Here, $\mathrm{sym}(\mX) = \frac{1}{2}(\mX + \mX^T)$, 
$\Exp(\cdot)$ denotes the matrix exponential and $Z =(\mY\mX^{-{1}})^{\frac{1}{2}}$.}
\label{tab:riemannian_tools}
\end{table}


\subsection{Learning ABLDs via Block Coordinate Descent}
We propose to use a block-coordinate descent (BCD) scheme to optimize our objectives in~\eqref{eq:iddl} and~\eqref{eq:ab_clustering}. In this scheme, each variable is updated alternately, while fixing others. We use the Riemannian conjugate gradient (RCG) algorithm~\cite{absil2009optimization} for optimizing over the dictionary atoms in IDDL (and cluster centroids in IDC). As our objective is non-convex in its variables (except for $\mW$), convergence of BCD iterations to a global minima is not guaranteed. In Alg.~\ref{alg:1} and Alg.~\ref{alg:2}, we detail out the meta-steps in our optimization scheme for solving our IDDL objective, while Alg.~\ref{alg:3} details the steps for optimizing our IDC objective. We initialize the dictionary atoms and the divergence parameters as described in Section~\ref{sec:param_init}. 

Recall from Section~\ref{sec:rcg} that an essential ingredient in RCG is efficient computations of the Euclidean gradients of the objective with respect to the variables. In the following, we derive expressions for these gradients. Note that we assume that the dictionary atoms (i.e., $\mB_i$) to be on an SPD manifold. 
As the optimization setup when both $\valpha,\vbeta$ are either positive or negative is essentially the same, in the derivations to follow, we assume $\valpha$ and $\vbeta$ belong to the non-negative orthant of the Euclidean space. We use the spectral projected gradient (SPG) descent algorithm ~\cite{birgin2001algorithm} for learning $\valpha$ and $\vbeta$. This algorithm uses Barzillai-Borwein step-size selection in the gradient descent, and projects the iterates into the non-negative orthant (or to the non-positive orthant if $\valpha,\vbeta<0$) to enforce the constraints. Empirically, using SPG is seen to converge very quickly, as is also observed in~\cite{cherian2016riemannian}.

\begin{algorithm} 
	\SetAlgoLined
	\KwIn{$\dataset$, $\mH$, $n$}
	$\dict \leftarrow \LEkmeans(\dataset, n)$, $(\valpha,\vbeta) \leftarrow \gridsearch$\;
	\Repeat{until convergence} 
	{		
		\For{$k=1$ \KwTo $n$}
		{
		    $\mB_k \leftarrow \text{update\_B}(\dataset, \mW, \valpha, \vbeta, \mB_k)$$\text{; // use \eqref{eq:update_bk_rr}}$
		}
		$(\valpha,\vbeta) \leftarrow \text{update\_$\alpha\beta$}(\dataset,\mW, \dict, \valpha, \vbeta)$$\text{; // use \eqref{eq:update_a_rr}}$\\
		$W \leftarrow \text{update\_$\mW$}$$\text{; // use~\eqref{eq:W}}$
	}
	\KwRet{$\dict,\valpha,\vbeta$}
	\caption{Block-Coordinate Descent for IDDL and Ridge Regression Loss.}
	\label{alg:1}
\end{algorithm}

\begin{algorithm} 
	\SetAlgoLined
	\KwIn{$\dataset$, $\mH$, $n$}
	$\dict \leftarrow \LEkmeans(\dataset, n)$, $(\valpha,\vbeta) \leftarrow \gridsearch$\;
	\Repeat{until convergence} 
	{		
		\For{$k=1$ \KwTo $n$}
		{
		    $\mB_k \leftarrow \text{update\_B}(\dataset, \mW, \valpha, \vbeta, \mB_k)$$\text{; // use \eqref{eq:update_bk_svm}}$
		}
		$(\valpha,\vbeta) \leftarrow \text{update\_$\alpha\beta$}(\dataset,\mW, \dict, \valpha, \vbeta)$$\text{; // use \eqref{eq:update_a_svm}}$\\
		\Repeat{until convergence}{
		$W \leftarrow \text{update\_$\mW$}$$\text{; // use~\eqref{eq:update_svm_1} and~\eqref{eq:update_svm_2}}$
		}
	}
	\KwRet{$\dict,\valpha,\vbeta$}
	\caption{Block-Coordinate Descent for IDDL Structured SVM Loss.}
    \label{alg:2}
\end{algorithm}


\begin{algorithm} 
	\SetAlgoLined
	\KwIn{$\dataset$, $k$}
	$\mathbf{C} \leftarrow \LEkmeans(\dataset, n)$, $(\alpha,\beta) \leftarrow \text{init}(lb, up)$\;
	\Repeat{until convergence} 
	{   
	\For{$z=1$ \KwTo $k$}
		{
		    $\mC_z \leftarrow \text{update\_C}(\dataset, \mathbf{\Pi}, \alpha, \beta, \mC_z)$$\text{; // use \eqref{eq:gradB_simplified}}$
		}
	
	    $(\alpha,\beta) \leftarrow \text{update\_$\alpha\beta$}(\dataset, \mathbf{\Pi}, \alpha, \beta, \mC_z)$$\text{; // use~\eqref{eq:abld_grad_a}}$
	    $\mathbf{\Pi} \leftarrow \text{update\_}\Pi(\dataset, \alpha, \beta, \mC)$$\text{; // use~\eqref{eq:update_partition}}$\\
		
	}
	\KwRet{$\mC, \mathbf{\Pi},\alpha,\beta$}
	\caption{Overview of Block-Coordinate Descent for $\idc$.}
	\label{alg:3}
\end{algorithm}

\subsection{Gradients on ABLD}
To complete the optimization schemes presented above, we need the gradients of our objectives with respect to the model parameters, which we present now. First, towards optimizing for the parameters of the divergence, quantities of particular interest are the derivatives of the ABLD with respect to its parameters $\alpha$ and $\beta$. Second, an equally important quantity for our learning schemes is the derivative of the ABLD with respect to the dictionary atoms. Furthermore, we explicitly handle the limiting case for $\alpha=\beta\rightarrow 0$ via the explicit derivation of the derivative of the ABLD with respect to its input matrices.

\subsubsection{\textbf{Derivative of ABLD wrt $\alpha$}}
Towards computing the derivative of ABLD with respect to its parameter $\alpha$, we use the form of the ABLD given in~\eqref{eq:abld_lambda} that involves the generalized eigenvalues of $\mX\inv{\mY}$. Letting $\theta = \alpha + \beta$ and $\nu = \alpha\beta$, for $\mX, \mY \in \spd{d}$ the derivative has the form:

\begin{align}
    &\nabla_{\alpha} D^{(\alpha,\beta)}(\mX  \parallel \mY) = \sum_{i=1}^d \nabla_{\alpha} \left[\frac{1}{\nu} \log\frac{\salpha\lambda_{i}^{\sbeta} + \sbeta\lambda_i^{-\salpha}}{\theta}\right] \notag \\
    &=\frac{1}{\salpha\nu}\sum_{i=1}^d \Bigg\{
    \frac{\salpha\lambda_{i}^{\sbeta} - \nu\lambda_{i}^{-\alpha}\log{\lambda_{i}}}{\alpha\lambda_{i}^{\beta} + \beta\lambda_{i}^{-\alpha}} -\frac{\alpha}{\theta} -\log{\frac{\alpha\lambda_{i}^{\beta} + \beta\lambda_{i}^{-\alpha}}{\theta}} \Bigg\}.
    \label{eq:abld_grad_a}
\end{align}
Using the dual symmetry property of ABLD reviewed in~\eqref{eq:dual_sym}, derivative of ABLD with respect to $\beta$ is straightforward.

\subsubsection{\textbf{Derivative of ABLD wrt $\mY$}}
\label{sec:dict_learn}
Towards deriving the derivative of the ABLD with respect to its input matrix $\mY$, we use the form of the divergence given in~\eqref{eq:abld}. In that way, letting $\rho = \frac{\alpha}{\beta}$ and $\mZ = \inv{\mX}$, the derivative of~\eqref{eq:abld} with respect to matrix $\mY$ has the form:

\begin{align}
    \nabla_{\mY} D^{(\alpha,\beta)}(\mX  \parallel \mY) &= \!\frac{1}{\nu}\!\nabla_{\mY}\!\! \logdet\left[\rho\left(\mZ\mY\right)^{\theta} + \eye{d}\right] +\frac{1}{\beta}\inv{\mY}.
    \label{eq:15}
\end{align}

The following theorem will come handy when we design gradients in our dictionary learning setup.
\begin{theorem}
\label{theor:logdet_der}
For $\mA,\mB \in \spd{d}$ and $p,q \geq 0$, 
\begin{align*}
\grad{B}&\logdet \left[p \left(\mA\mB\right)^{q} + \eye{d}\right] =\\ &pq\inv{\mB}\mA^{-\half}\!\!\left(\mA^{\half}\mB\mA^{\half}\right)^{q} 
\times\left(\eye{d} + p \left(\mA^{\half}\mB \mA^{\half}\right)^{q}\right)^{\!-1}\hspace*{-0.3cm}\mA^{\half}.
\end{align*}
\end{theorem}

Making use of Theorem~\ref{theor:logdet_der}, the derivative described in~\eqref{eq:15} becomes:

\begin{align}
\nabla_{\mY} D^{(\alpha,\beta)}&(\mX  \parallel \mY) = \rho\theta \inv{\mY}\mZ^{-\half}\!\!\left(\mZ^{\half}\mY\mZ^{\half}\right)^{\!\!\theta} \notag \\
&\times \left(\eye{d} + \rho \left(\mZ^{\half}\mY \mZ^{\half}\right)^{\theta}\right)^{\!-1}\mZ^{\half} +\frac{1}{\beta}\inv{\mY}.
\label{eq:21}
\end{align}

\begin{remark}
Computing~\eqref{eq:21} for large datasets can become overwhelming. A more efficient implementation could be the following. Say, $(\mU,\mDelta)$ be the Schur decomposition of $\mZ^{\half}\mY\mZ^{\half}$, which is faster to compute than the eigenvalue decomposition~\cite{golub2012matrix} (required for computing $\mZ^{\half}$). With $\vdelta=\diag(\mDelta)$,~\eqref{eq:21} can be rewritten as:
\begin{align}
\nabla_{\mY} D^{(\alpha,\beta)}(\mX  \parallel \mY) &= \rho\theta\inv{\mY}\left(\mZ^{-\half}\mU\right) \left[\diag\left(\frac{\vdelta^\theta}{1+\rho\vdelta^{\theta}}\right)\right] \notag \\ 
& \times \inv{\left(\mZ^{-\half}\mU\right)} +\frac{1}{\beta}\inv{\mY}.\label{eq:gradB_simplified}
\end{align}
Compared to~\eqref{eq:21}, this reduces the number of matrix multiplications from 5 to 3 and matrix inversions from 2 to 1.
\end{remark}

\subsubsection{\textbf{Derivative of the ABLD for $\alpha=\beta\rightarrow 0$ wrt $\mY$}}
As alluded to earlier, ABLD is non-smooth at the origin and we need to resort to the limit of the divergence, which happens to be the natural Riemannian metric (AIRM). That is,
\begin{equation}
    \abldab{\mX}{\mY}{0}{0} = \fnorm{\Log{\Big( \mX^{-\half}\mY\mX^{-\half}\Big)}}^2.
    \label{def_AIRM}
\end{equation}
\noindent Letting $P = \mX^{-\half}\mY\mX^{-\half}$, the derivative of~\eqref{def_AIRM} with respect to matrix $\mY$ is given by:
\begin{equation}
    \nabla_{\mY} \abldab{\mX}{\mY}{0}{0} = 2 \mX^{-\half} \left(\Log{P}\right)\inv{P} \mX^{-\half}.
    \label{eq:grad_airm}
\end{equation}
\noindent Note that a simplification similar to~\eqref{eq:gradB_simplified} is also possible for~\eqref{eq:grad_airm}.

With this general gradients for learning $\alpha,\beta$, dictionary and the cluster centroids, now we consider learning parameters specific to each of our objectives.

\subsection{Optimizing IDDL -- Ridge Regression Objective}
Reconsidering our ridge regression loss $f_1$ defined in~\eqref{eq:ridge}, suppose  $\mV$ and $\mH$ are matrices obtained by stacking $\vv_i$ and $\vh_i$ along their $i$-th column, for $i=1,2,\cdots, N$. When fixing $\dict,\valpha$ and $\vbeta$, the objective can be solved in closed form as:
\begin{equation}
    \mW^* = \mH\mV^T\inv{(\mV\mV^T + \gamma\eye{d})},
    \label{eq:W}
\end{equation}

Towards deriving the gradient of the IDDL objective for $f_1$ w.r.t. the $k^{th}$ dictionary atom $\mB_k$ we capitalize on the observation that only the $k$-th dimension of $\vv_i$ involves $\mB_k$. To simplify the notation, let us assume:
\begin{equation}
    \zeta_i = -(\vh_i - \mW\vv_i)^T\mW,
    \label{eq:zeta}
\end{equation}
and let $\zeta_i^k$ be its $k$-th dimension. Then we have:
\begin{align}
\nabla_{\mB_k} f_1 = \zeta^k_i \nabla_{\mB_k}\left(\abldab{\mX_i}{\mB_k}{\valpha^k}{\vbeta^k}\right),
\label{eq:update_bk_rr}
\end{align}

\noindent where the gradient of the ABLD w.r.t. the $k$-th atom is derived in \eqref{eq:gradB_simplified}. Similarly, for gradients w.r.t. $\valpha_k$, using the derivative of the ABLD derived in~\eqref{eq:abld_grad_a}, we get:

\begin{align}
    \nabla_{\valpha_k} f_1 &= \zeta^k_i \nabla_{\valpha_k}\left(\abldab{\mX_i}{\mB_k}{\valpha^k}{\vbeta^k}\right).
    \label{eq:update_a_rr}
\end{align}

It should be noted that the gradients w.r.t. $\vbeta_k$ can be easily computed using the dual symmetry property described in~\eqref{eq:dual_sym}.

\subsection{Optimizing IDDL -- Structured SVM Objective}
\label{sec:ssvm}
Looking back at our SVM loss $f_2$ in~\eqref{eq:svm}; differentiating $f_2$ w.r.t. the rows of $W$ while fixing $\dict,\valpha$, and $\vbeta$ we get:

\begin{align}
    \nabla_{\mathbf{w}_{y_i}} f_2 = - \left( \sum_{j \neq y_i} \mathbbm{1}_{\left( \mathbf{w}_j^T \mathbf{v}_i - \mathbf{w}_{y_i}^T \mathbf{v}_i + \Delta > 0 \right)}\right) \mathbf{v}_i + 2\gamma \mathbf{w}_{y_i}
    \label{eq:update_svm_1}
\end{align}  

\begin{align}
    \nabla_{w_{j \neq y_i}} f_2 = \mathbbm{1}_{\left( \mathbf{w}_j^T \mathbf{v}_i - \mathbf{w}_{y_i}^T \mathbf{v}_i + \Delta > 0 \right)} \mathbf{v}_i + 2\gamma \mathbf{w}_j
    \label{eq:update_svm_2}
\end{align}

\noindent where $\mathbbm{1}_{(.)}$ is the indicator function. Similarly to our derivations for $f_1$, to simplify our notations we let:

\begin{align}
    \xi_i = - \sum_{j \neq y_i} \mathbbm{1}_{\left( \mathbf{w}_j^T \mathbf{v}_i - \mathbf{w}_{y_i}^T \mathbf{v}_i + \Delta > 0 \right)} \left( \mathbf{w}_j - \mathbf{w}_{y_i}\right).
\end{align}

\noindent and let $\xi_i^k$ be its $k$-th dimension. That way, making use of~\eqref{eq:gradB_simplified} and~\eqref{eq:abld_grad_a}, we define the gradients of $f_2$ w.r.t to $\mB_k$ and $\valpha_k$, respectively as:

\begin{align}
\nabla_{\mB_k} f_2 = \xi^k_i \nabla_{\mB_k}\left(\abldab{\mX_i}{\mB_k}{\valpha^k}{\vbeta^k}\right),
\label{eq:update_bk_svm}
\end{align}

\begin{align}
    \nabla_{\valpha_k} f_2 &= \xi^k_i \nabla_{\valpha_k}\left(\abldab{\mX_i}{\mB_k}{\valpha^k}{\vbeta^k}\right).
    \label{eq:update_a_svm}
\end{align}

\subsection{Optimizing IDC -- Clustering Objective}
The gradients of $f_3$ with respect to both the divergence parameters, as well as the clustering centroids, are obtained directly from~\eqref{eq:abld_grad_a} and~\eqref{eq:gradB_simplified}, respectively. Lastly, to update $\mathbf{\Pi}$ in~\eqref{eq:ab_clustering}, we need to find the cluster centroid $\mC_\pi$ nearest to a given data point $\mX_i$, for which we solve the following argmin problem, by assuming the ABLD parameters are fixed at the current iterate. Formally, the data points in the cluster $\pi_z$ are updated as $\pi_{z^*}\rightarrow \pi_{z^*} \cup \set{\mX_i}$, where:

\begin{align}
z^* = \argmin_{\forall z\in\set{1,2,\cdots, k}} \abldab{\mX_i}{\mathbf{C}_z}{\alpha}{\beta}.
\label{eq:update_partition}
\end{align}

\subsection{Computational Complexity}
It should be noted that terms such as $\inv{\mX_i}$, which can be computed offline are omitted from this analysis. Using the simplifications depicted in~\eqref{eq:gradB_simplified} and Schur decomposition, gradient computation for each dictionary atom or centroid takes $\bigoh(Nd^3)$ flops. Using the gradient formulation in~\eqref{eq:abld_grad_a} for $\valpha$ and $\vbeta$, we need $\bigoh(Ndn + Nd^3)$ flops. Computations of the closed form for $\mW$ using the ridge regression loss in~\eqref{eq:W} takes $\bigoh(n^2(L+N)+n^3+nLN)$. For the discriminative setup, at test time, given that we have learned the dictionary and the parameters of the divergence, encoding a data matrix requires $\bigoh(nd^3)$ flops, which is similar in complexity to the recent sparse coding schemes such as~\cite{cherian2016riemannian}. As for the gradient computation for each $\mC$ in IDC takes $\bigoh(Nd^3)$ flops and the overall clustering setup takes $\bigoh(Ndk + Nd^3)$ flops, similar in complexity to a Karcher mean algorithm~\cite{bini2013computing} using AIRM as the similarity measure. 
\section{Experiments}
\label{sec:expts}
In this section, we present a thorough evaluation of the proposed inference schemes on a diverse set of computer vision datasets. We use the following eight datasets, namely (i) JHMDB~\cite{jhuang2013towards}, (ii) HMDB~\cite{kuehne2011hmdb}  (iii) KTH-TIPS2~\cite{mallikarjuna2006kth}, (iv) Brodatz textures~\cite{ojala1996comparative}, (v) Virus~\cite{kylberg2012segmentation}, (vi) SHREC 3D~\cite{lai2011large}, (vii) Myometrium cancer~\cite{panos_icpr}, and (viii) Breast cancer~\cite{panos_icpr}. Below, we provide details of all these datasets, and how SPD descriptors are obtained on them. 

We use standard evaluation schemes reported previously on these datasets. In some cases, we use our own implementations of popular methods but strictly following the recommended evaluation settings. For those datasets that do not have prescribed cross-validation splits, we repeat the experiments at least 5 times and average the performance scores. For our SVM-based experiments, we use a linear SVM on the log-Euclidean mapped SPD matrices.

\subsection{Datasets}
\noindent\textbf{HMDB~\cite{kuehne2011hmdb} and JHMDB~\cite{jhuang2013towards} datasets}: These are two popular action recognition benchmarks. The HMDB dataset consists of 51 action classes associated with 6766 video sequences, each sequence with 30--400 video frames. JHMDB is a subset of HMDB with 955 sequences in 21 action classes with 15--40 frames in each video, where the video subset in JHMDB has human pose better visible. Following the official train/test split, we use 70\% of the videos for training and rest for testing on both datasets. To generate SPD matrices on these datasets, we use the scheme proposed in~\cite{cherian_wacv}, where we compute RBF kernel descriptors on the output of per-frame CNN class predictions (fc8) for each stream (RBF and optical flow) separately, and fusing these two SPD matrices into a single block-diagonal matrix per sequence. For the two-stream model, we use a VGG16 model trained on optical flow and RGB frames separately as described in~\cite{simonyan2014two}. Thus, our descriptors are of size $102\times 102$ for HMDB and $42\times 42$ for JHMDB.

\noindent\textbf{SHREC 3D Object Recognition Dataset~\cite{lai2011large}:} It consists of 15000 RGBD covariance descriptors generated from the SHREC dataset~\cite{lai2011large} by following~\cite{fehr2013covariance}. SHREC consists of 51 3D object classes. The descriptors are of size $18\times 18$. Similar to~\cite{cherian2016riemannian}, we randomly picked 80\% of the dataset for training and used the remaining for testing.

\noindent\textbf{KTH-TIPS2 dataset~\cite{mallikarjuna2006kth} and Brodatz Textures~\cite{ojala1996comparative}:} These are popular texture recognition datasets. The KTH-TIPS dataset consists of 4752 images from 11 material classes under varying conditions of illumination, pose, and scale. Covariance descriptors of size $23\times 23$ are generated from this dataset following the procedure in~\cite{harandi2014bregman}. We use the standard 4-split cross-validation for our evaluations on this dataset. As for the Brodatz dataset, we used 100 texture images for our experiments, each image is $640\times 640$ resoultion. To produce the covariance descriptors, we follow the procedure outlined in~\cite{cherian2016riemannian}. Specifically, we extracted $32\times 32$ non-overlapping patches from each image. Next, to produce the covariance descriptor for a given patch, we used the relative pixel coordinates for all pixels in the patch, its image intensity, and respective image gradients, which form 5-dimensional features, from which $5\times 5$ region covariance descriptors are produced for the respective patch. Our dataset consists of 31000 SPD matrices. As proposed in~\cite{cherian2016riemannian}, for our evaluation use an 80:20 rule as in the RGBD dataset above.

\noindent\textbf{Virus Dataset~\cite{kylberg2012segmentation}:}  It consists of 1500 images of 15 different virus types. Similar to the KTH-TIPS, we use the procedure in~\cite{harandi2014bregman} to generate $29\times 29$ covariance descriptors from this dataset and follow their evaluation scheme using three-splits.

\noindent\textbf{Cancer Datasets~\cite{panos_icpr}.} Apart from these standard SPD datasets, we also report performances on two cancer recognition datasets from~\cite{panos_icpr}. We use images from two types of cancers, namely (i) Breast cancer, consisting of binary classes (tissue is either cancerous or not) consisting of about 3500 samples, and (ii) Myometrium cancer, consisting of 3320 samples; we use covariance-kernel descriptors as described in~\cite{panos_icpr} which are of size $8\times 8$. We follow the 80:20 rule for evaluation on this dataset as well.

\begin{figure*}[htbp]
\centering
\subfigure[JHMDB]{\label{fig:jhmdb}\includegraphics[width=5cm, trim={0cm 0cm 0cm 0cm},clip]{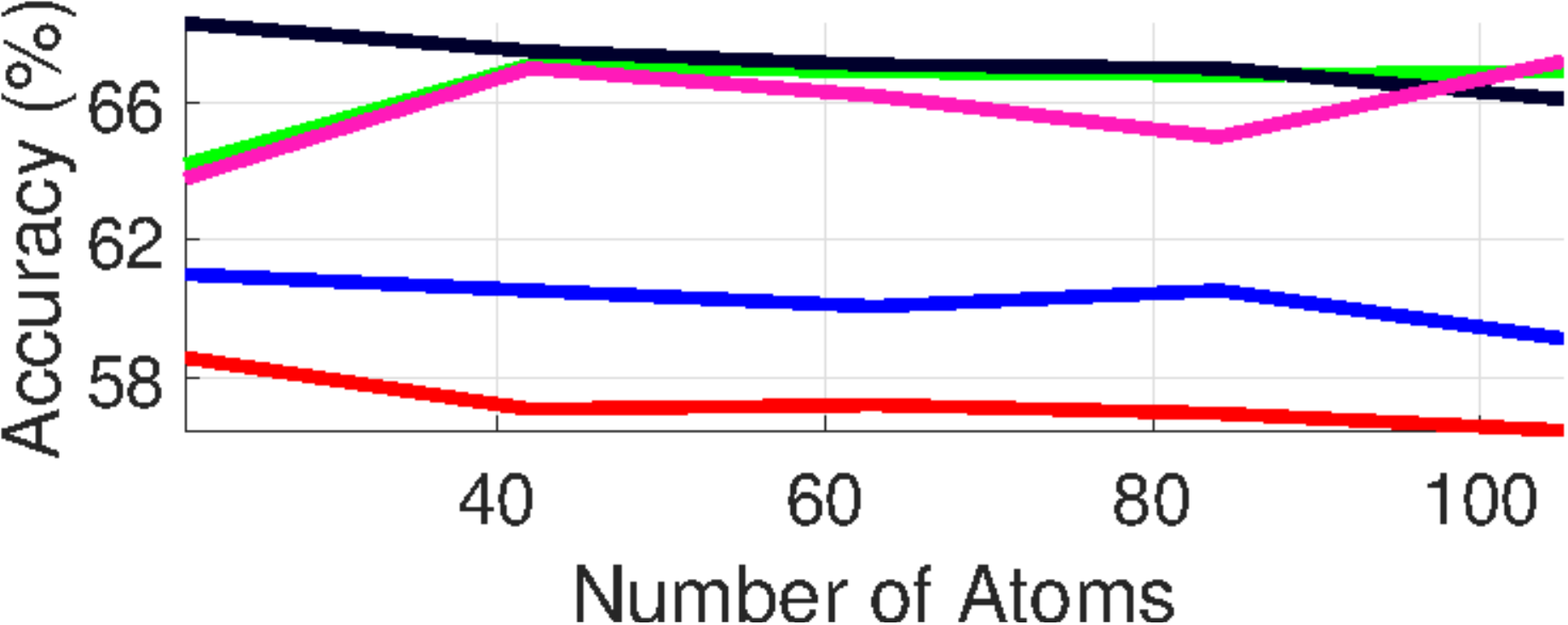}}\hspace*{0.2cm}
\subfigure[VIRUS]{\label{fig:virus}\includegraphics[width=5cm, trim={0cm 0cm 0cm 0cm},clip]{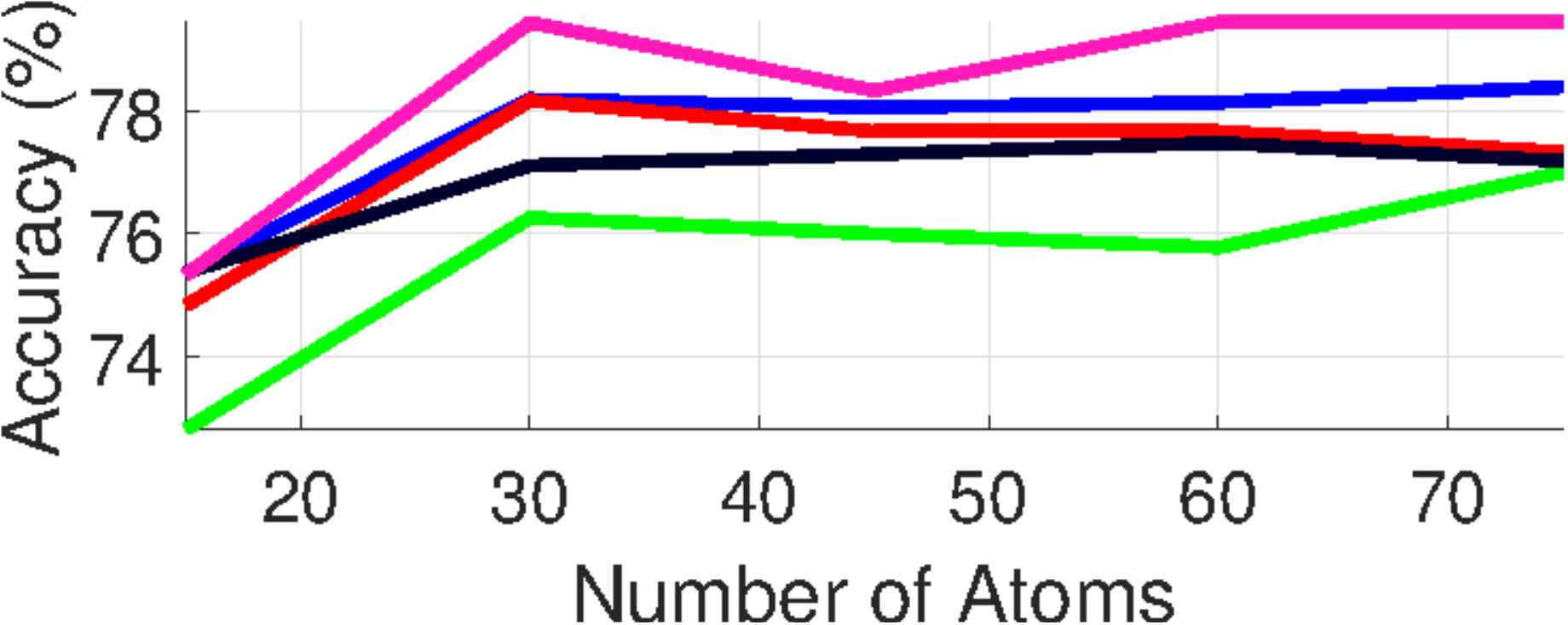}}\hspace*{0.2cm}
\subfigure[KTH-TIPS2]{\label{fig:kth}\includegraphics[width=5.7cm, trim={0cm 0cm 0cm 0cm},clip]{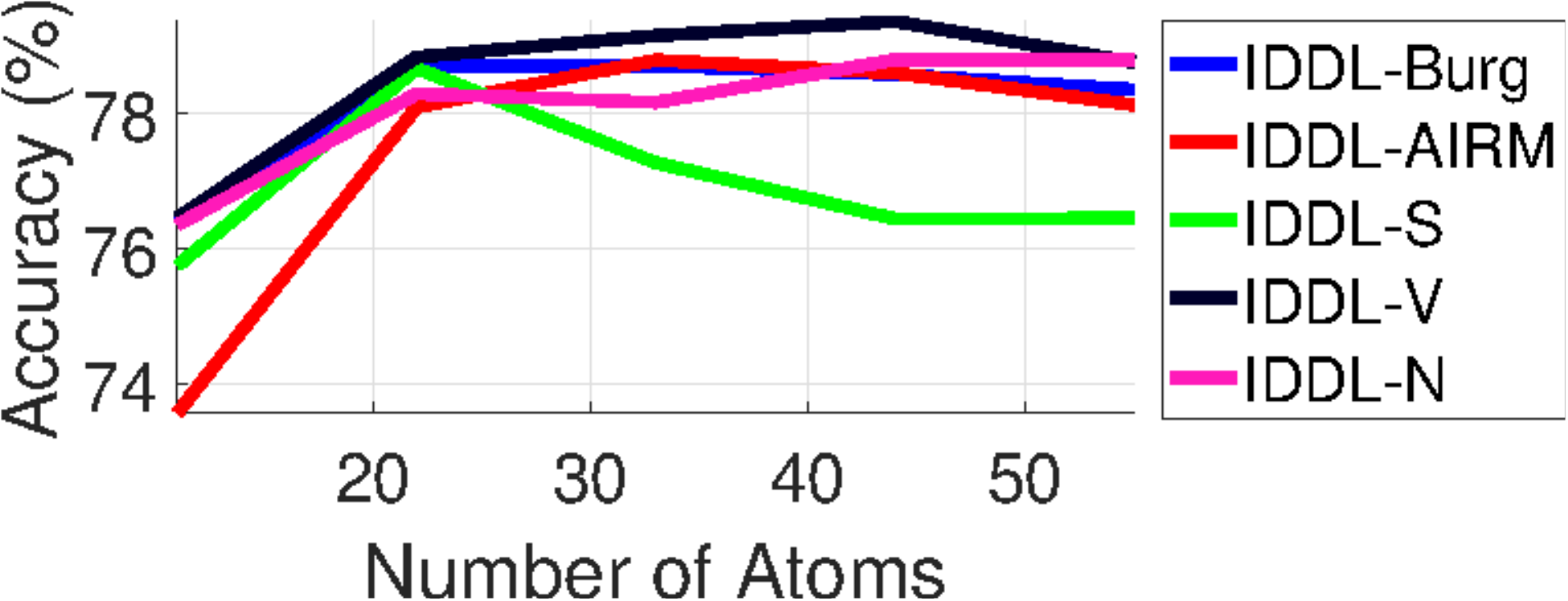}}\\
\subfigure[RGB-D Objects]{\label{fig:duc}\includegraphics[width=5cm, trim={0cm 0cm 0cm 0cm},clip]{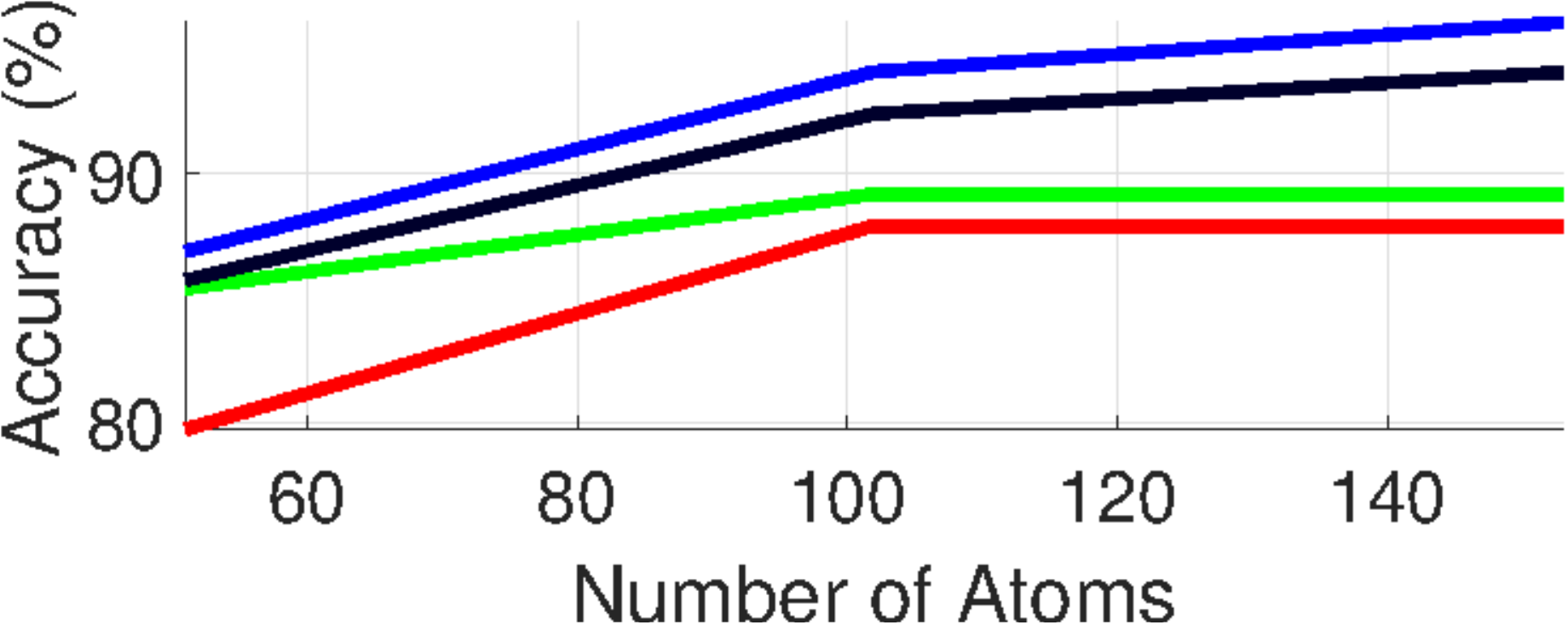}}\hspace*{0.2cm}
\subfigure[Breast Cancer]{\label{fig:bcancer}\includegraphics[width=5cm, trim={0cm 0cm 0cm 0cm},clip]{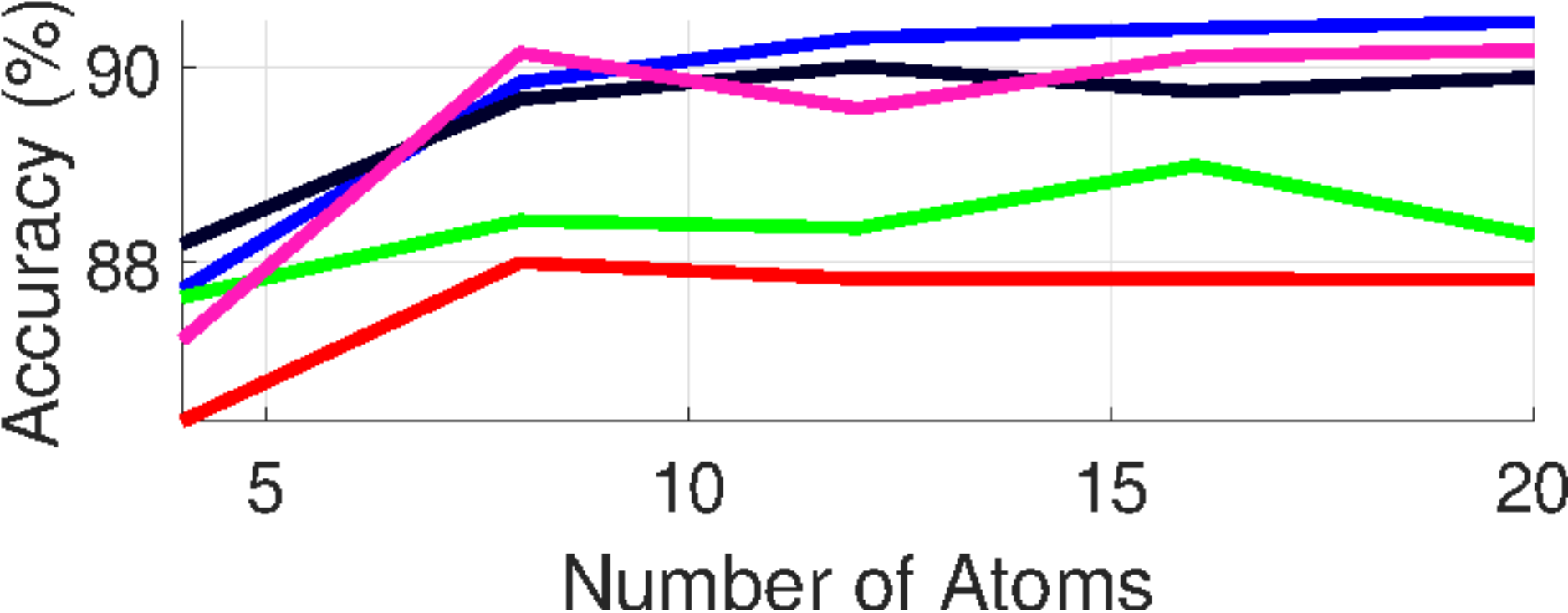}}\hspace*{0.2cm}
\subfigure[Myometrium Cancer]{\label{fig:mcancer}\includegraphics[width=5.6cm, trim={0cm 0cm 0cm 0cm},clip]{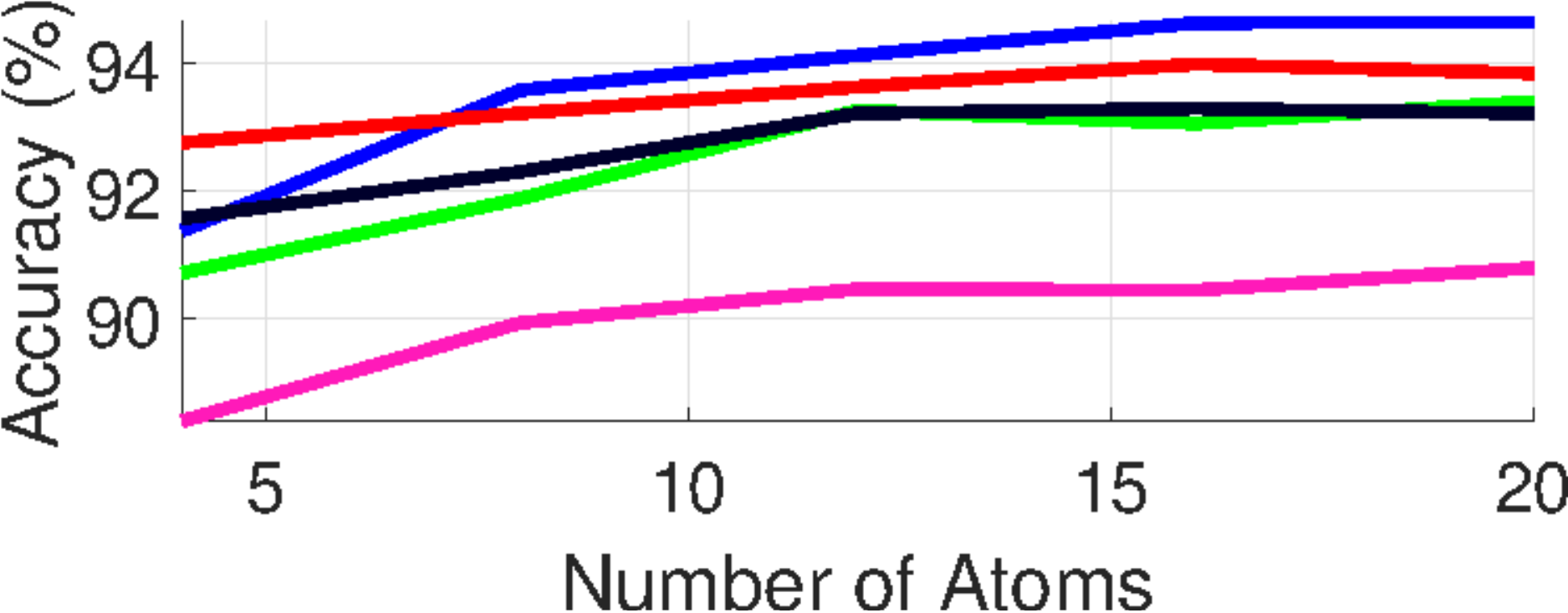}}
\caption{\label{fig:comp1st} \footnotesize{Comparisons between different variants of IDDL for increasing number of dictionary atoms.}}
\label{fig:illustrative_examples}
\end{figure*}

\subsection{Experimental Setup}
Since we present experiments on a variety of datasets and under various configurations, we summarize our main experiments first. There are four sets of experiments we conduct, namely (i) an ablative study of various parameters, learning configurations, convergence, and run-time analysis of our problem setup, (ii) comparison of IDDL against other popular measures on SPD matrices, (iii) comparisons among various configurations of IDDL, and (iv) comparisons against state of the art approaches on the above datasets. We follow a similar trend in our experimental setup for our clustering setup.  


\subsection{Parameter Initialization}
\label{sec:param_init}

In all the experiments, we initialized the parameters of IDDL (e.g., the initial dictionary) in a principle-way. We initialized the dictionary atoms by applying log-Euclidean K-Means; i.e., we compute the log-Euclidean map of the SPD data, compute Euclidean K-Means on these mapped points, and remap the K-Means centroids to the SPD manifold via an exponential map. To initialize $\alpha$ and $\beta$, we recommend grid-search by fixing the dictionary atoms as above. As an alternative to the grid-search, we empirically observed that a good choice is to start with the Burg divergence (i.e., $\alpha=\beta=1$).
The regularization parameter $\gamma$ in IDDL was chosen using cross-validation, while the regularization parameter $\mu$ for $\idc$ was set to 1.
\subsection{Ablative Study}
In this section, we study the influence of each of the components in our algorithm. First, we demonstrate how the performance on a dataset varies when changing the parameters $\alpha$ and $\beta$ in ABLD. This forms the basis of all our further experiments. 

\subsubsection{Performance for Varying $\alpha$, $\beta$}
\label{sec:varying_alphabeta}
\begin{figure*}[!htbp]
\centering
{\subfigure[]{\label{fig:heatmap-a}\includegraphics[width=4.4cm, trim={1cm 0cm 3cm 2.2cm},clip]{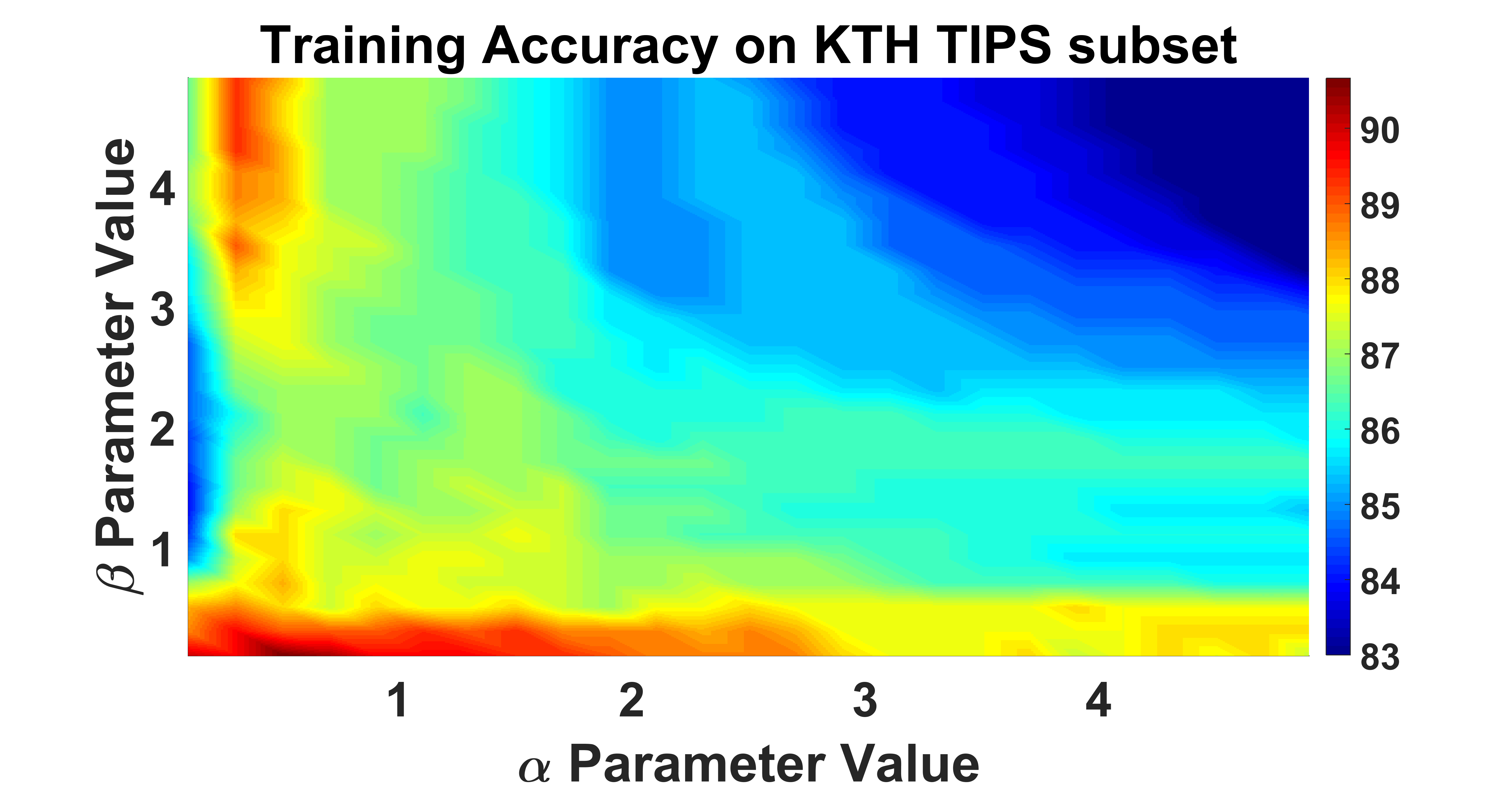}}}
{\subfigure[]{\label{fig:heatmap-b}\includegraphics[width=4.4cm, trim={1cm 0cm 3cm 2.2cm},clip]{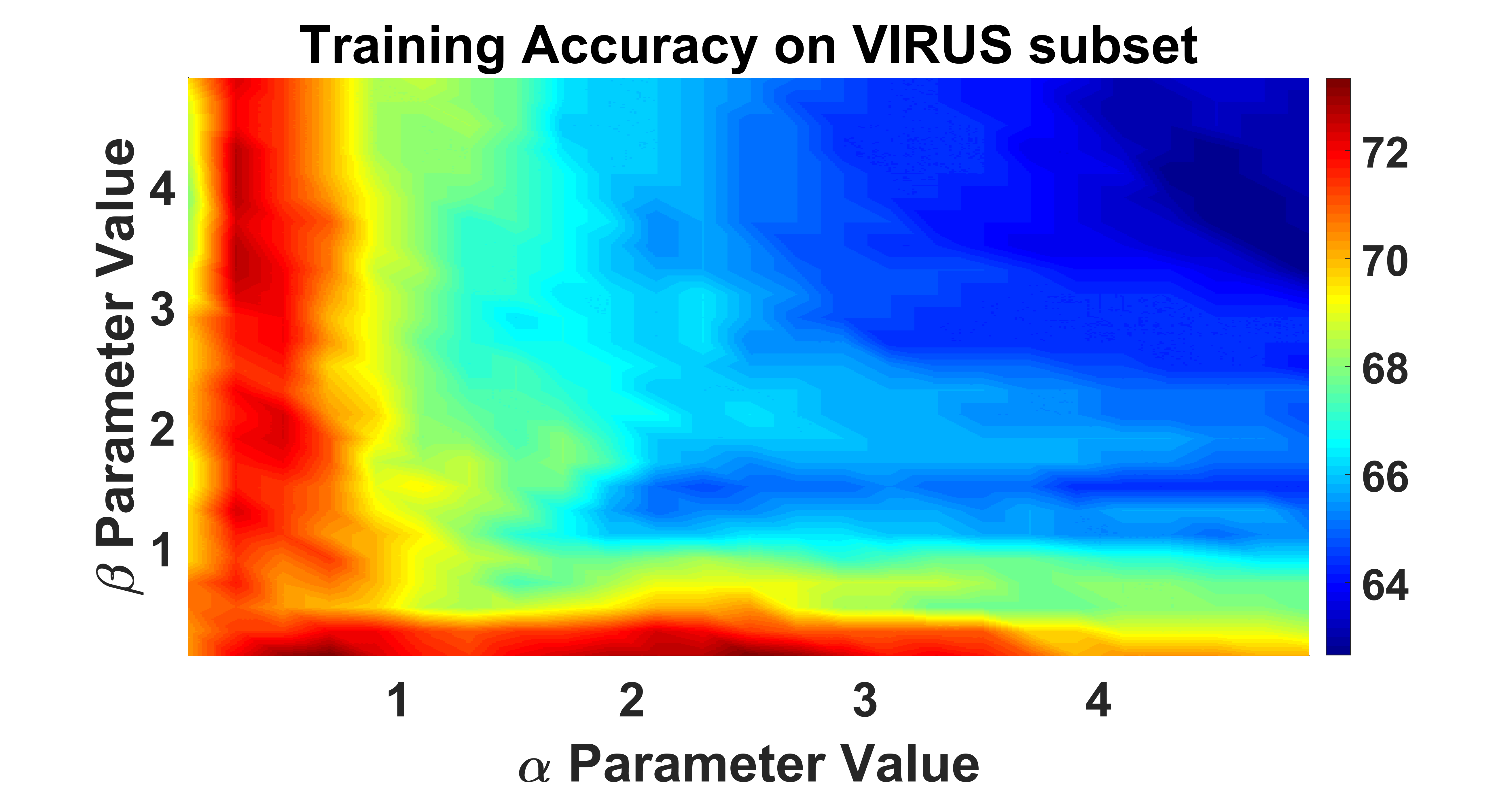}}}
\subfigure[]{\label{fig:conv-KTH}\includegraphics[width=4.5cm, trim={0cm 0cm 3cm 3.0cm},clip]{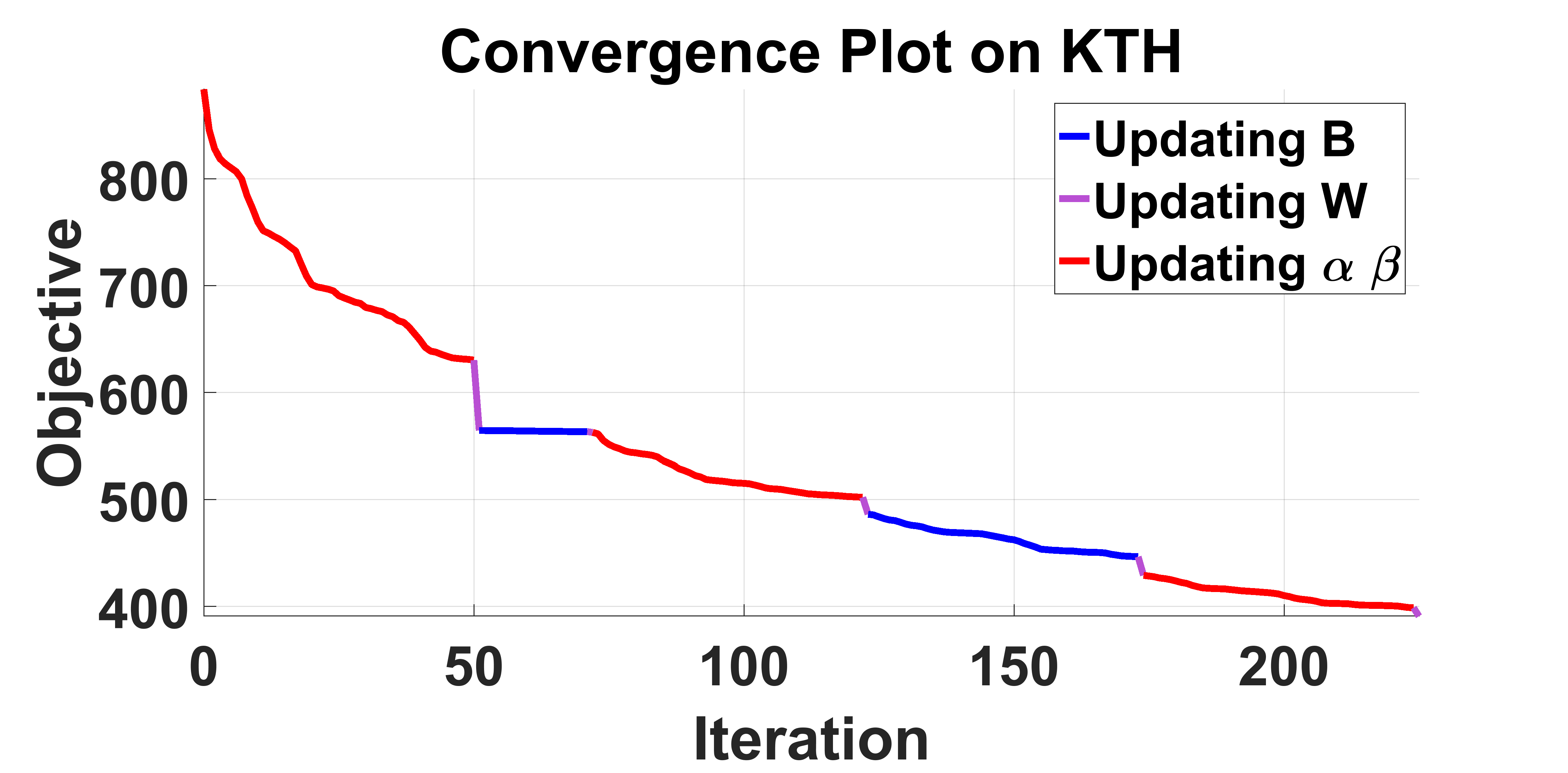}}
\subfigure[]{\label{fig:conv-virus}\includegraphics[width=4.5cm, trim={0cm 0cm 3cm 2.3cm},clip]{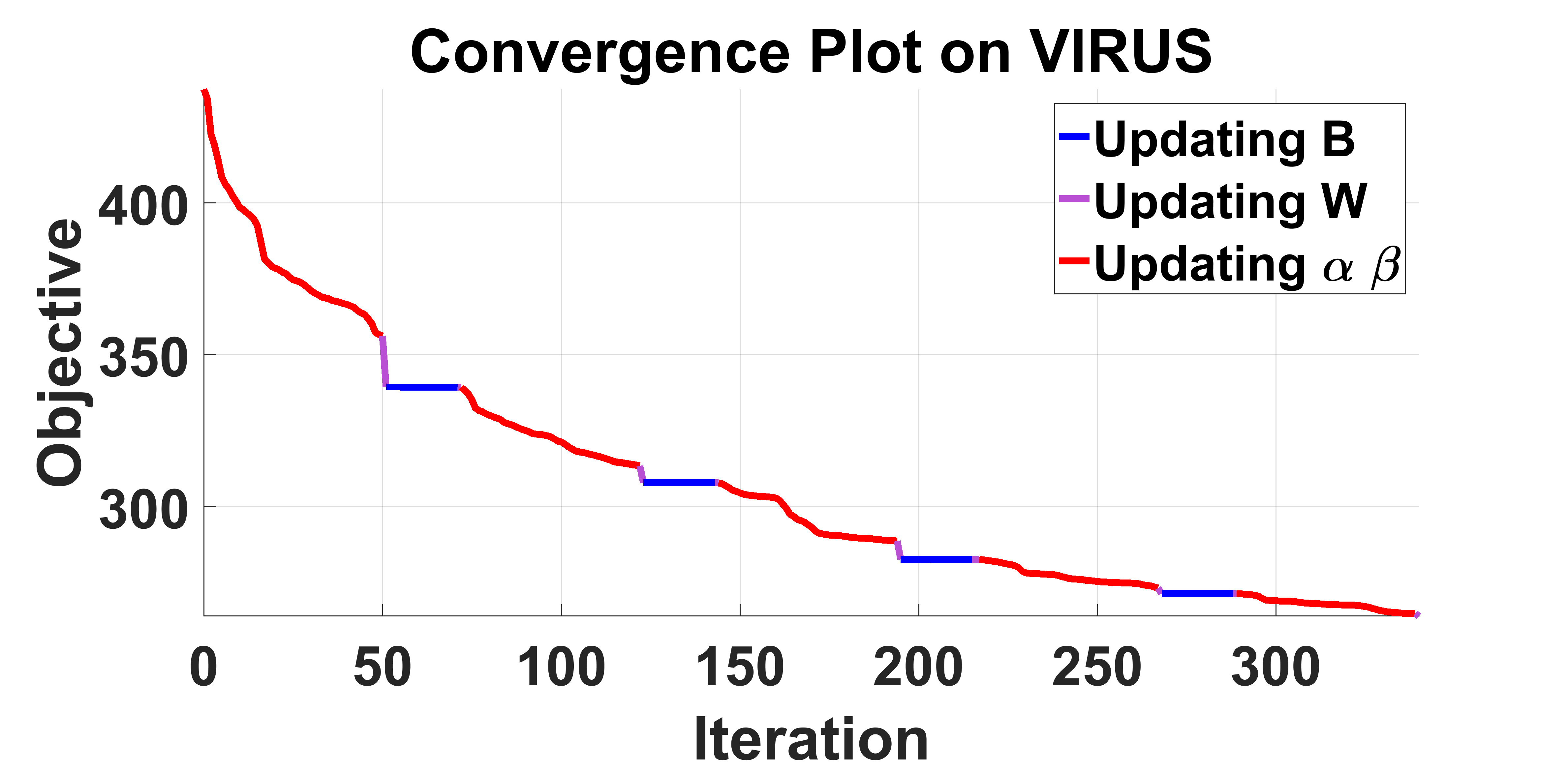}}
\caption{~\ref{fig:heatmap-a} and~\ref{fig:heatmap-b} show accuracy on KTH-TIPS2 and Virus datasets as we change $\alpha$ and $\beta$, fixing the number of dictionary atoms. See the algorithmic settings in Sec.~\ref{sec:varying_alphabeta}). \ref{fig:conv-KTH} and \ref{fig:conv-virus} show convergence of our optimization scheme for IDDL (ridge regression) on KTH-TIPS2 and Virus datasets. See the text for more details.
}
\end{figure*}

 In Figure~\ref{fig:heatmap-a} and~\ref{fig:heatmap-b}, we plot a heatmap of the classification accuracy against changing $\alpha$ and $\beta$ on the KTH-TIPS2 and Virus datasets. We fixed the size of dictionaries to 22 for the KTH TIPS and 30 for the Virus datasets. The plots  reveal that the performance varies for different parameter settings, thus (along with the results in Table~\ref{tab:1stlevel}) substantiates that learning these parameters is a way to improve performance. %

\subsubsection{Comparisons to Variants of IDDL}\label{sec:variants_IDDL}
In this section, we analyze various aspects of the performance of IDDL, using the ridge regression objective. Generally speaking, IDDL formulation is generic and customizable. For example, even though we formulated the problem as using a separate ABLD on each dictionary atom, it does not hurt to learn the same divergence over all atoms in some applications. To this end, we test the performance of three scenarios, namely (i) using a scalar $\alpha$ and $\beta$ that is shared across all the dictionary atoms (which we call IDDL-S), (ii) a vector $\alpha$ and $\beta$, where we assume $\alpha=\beta$, but each dictionary atom can potentially have a distinct parameter pair (we call this case IDDL-V), and (iii) the most generic case where we could have $\alpha$, $\beta$ as vectors and they may not be equal, which we refer as IDDL-N. 
In Figure~\ref{fig:comp1st}, we compare all these configurations on six of the datasets. We also include specific cases such as the Burg divergence ($\alpha=\beta=1$) and the AIRM case ($\alpha=\beta=0$) for comparisons (using the dictionary learning scheme proposed in Section~\ref{sec:dict_learn}). Our experiments show that IDDL-N and IDDL-V consistently perform well on almost all datasets. This is unsurprising given the generality of IDDL. While IDDL-S shows similar performance to other methods when the number of atoms is small, it drops for more number of atoms.

\subsubsection{Convergence Study}
\begin{figure}[!htbp]
\begin{center}
\includegraphics[width=4.4cm,trim={1cm 0cm 1cm 1cm},clip]{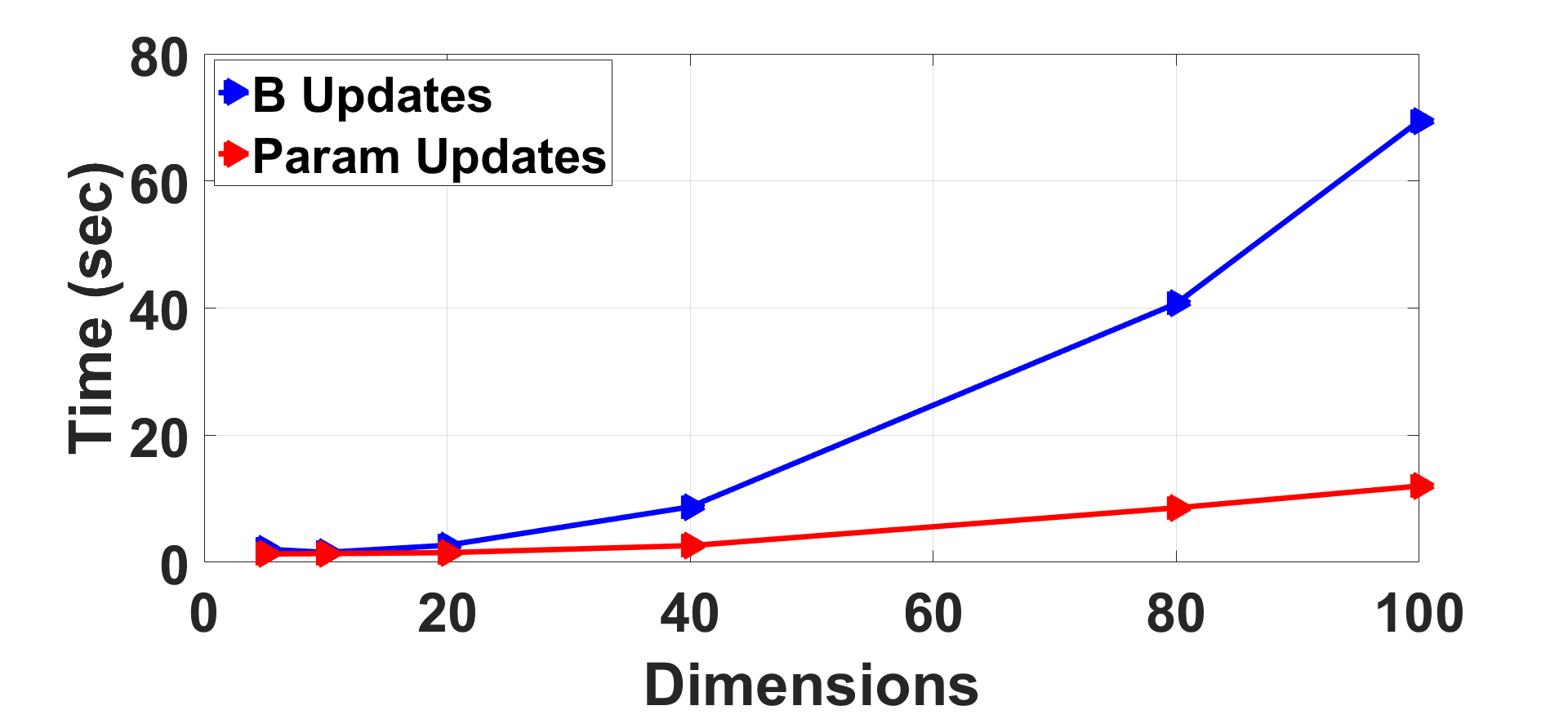}
\includegraphics[width=4.4cm,trim={1cm 0cm 1cm 1cm},clip]{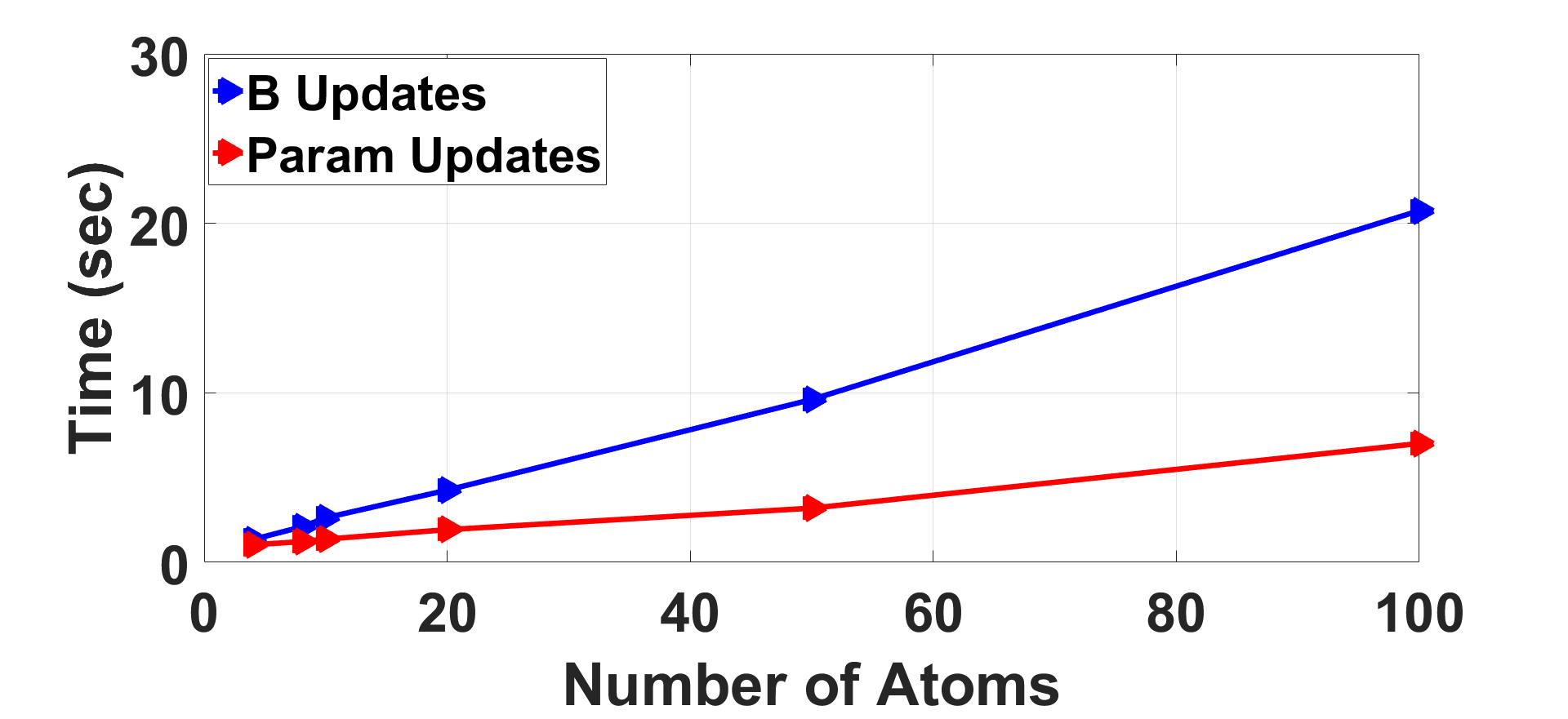}
\caption{\label{fig:timecomps} Time taken for one gradient computation and two objective functions evaluations against an increasing number of matrix dimensions (left) and number of dictionary atoms (right).}
\end{center}
\end{figure}

In Figures~\ref{fig:conv-KTH} and~\ref{fig:conv-virus}, we plot the convergence of our objective against iterations. We also depict the BCD objective as contributed by the dictionary learning updates and the parameter learning; we use the IDDL-V for this experiment. As is clear, most part of the decrement in objective happens when the dictionary is learned, which is not surprising given that it has the most number of variables to learn. For most datasets, we observe that the RCG converges in about 200-300 iterations. In Figure~\ref{fig:timecomps}, we plot the running time for one iteration of RCG against the number of dimensions of the matrices and the number of dictionary atoms. While our dictionary updates seem quadratic in the number of dimensions, it  scales linearly with the dictionary size.

\subsubsection{Evaluation of Joint Learning}
In Table~\ref{tab:paramatoms}, we evaluate the usefulness of learning the information divergence against learning the dictionary on the Virus dataset. For this experiment, we evaluated three scenarios, (i) fixing the dictionary to the initialization (using KMeans), and learning the parameters $\alpha,\beta$ using the IDDL-S variant, (ii) fixing $\alpha,\beta$ to the initialization using GridSearch, while learning the dictionary, and (iii) learning both dictionary and the parameters jointly. As the results in Table~\ref{tab:paramatoms} shows, jointly learning the parameters demonstrates better results, thus justifying our IDDL formulation.

\begin{table}[htbp]
\centering
\renewcommand{\arraystretch}{1.2} 
\begin{tabular}{c|c|c|c}
\textbf{Atoms | Method}   & \textbf{IDDL-Fix$(\alpha,\beta)$}  & \textbf{IDDL-Fix$(\mathbf{B})$} & \textbf{IDDL-N}\\
\hline
\textbf{15} & \textbf{78.33\%}& 61.67\% & 77.33\%\\
\hline
\textbf{45} & 80.33\%& 70.00\% & \textbf{83.67\%}\\
\hline
\textbf{75} & 81.67\%& 76.00\% & \textbf{82.33\%}\\
\end{tabular}
\caption{Performance evaluation of IDDL on a single split of the Virus dataset when fixing the dictionary atoms against fixing the parameters, and jointly learning the atoms and the parameters.}
\label{tab:paramatoms}
\end{table}

\subsubsection{Trajectories of $\alpha,\beta$}
\begin{figure*}[t]
\subfigure[]{\includegraphics[width=6cm,trim={0 7cm 1cm 7cm},clip]{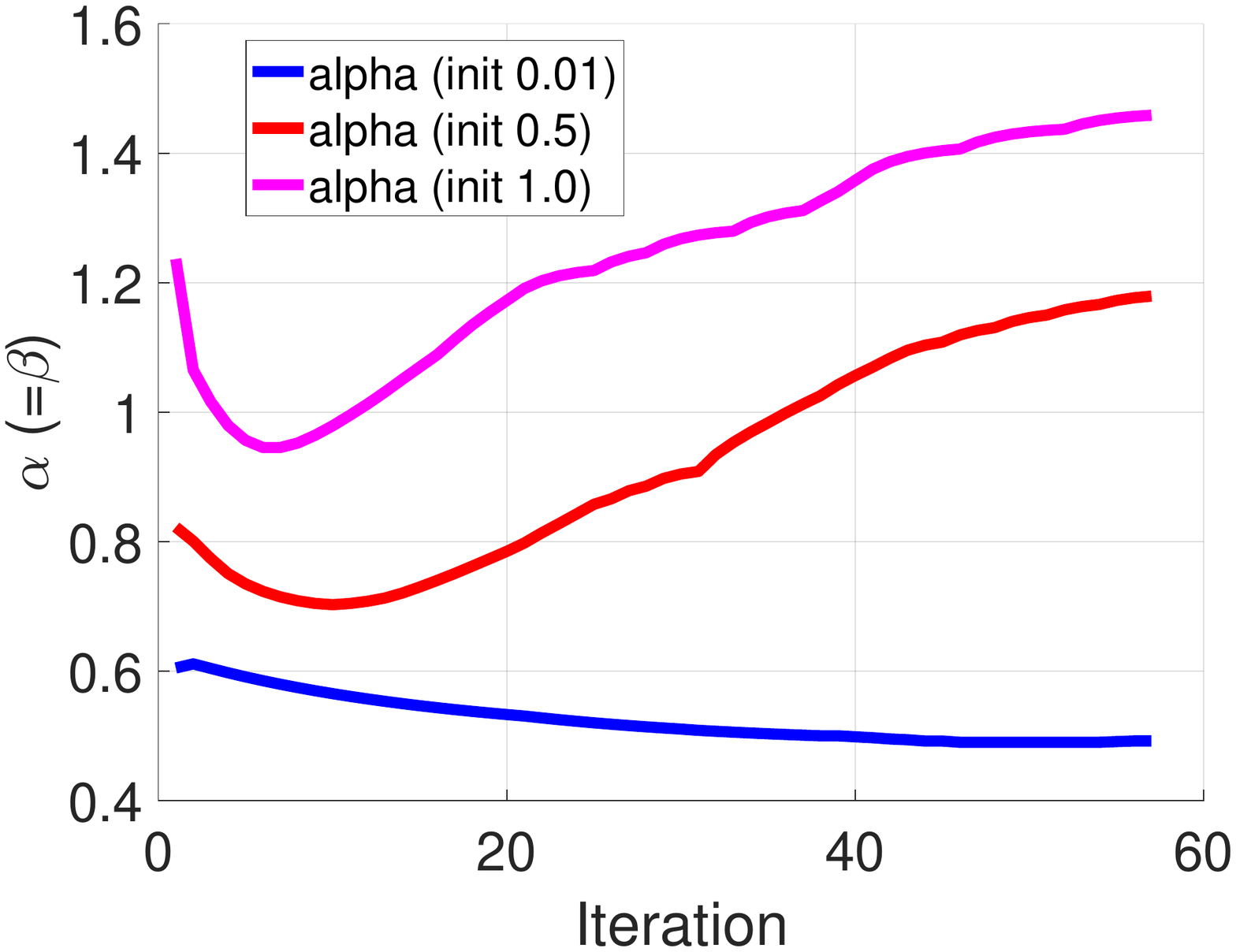}}
\subfigure[]{\includegraphics[width=6cm,trim={0 7cm 1cm 7cm},clip]{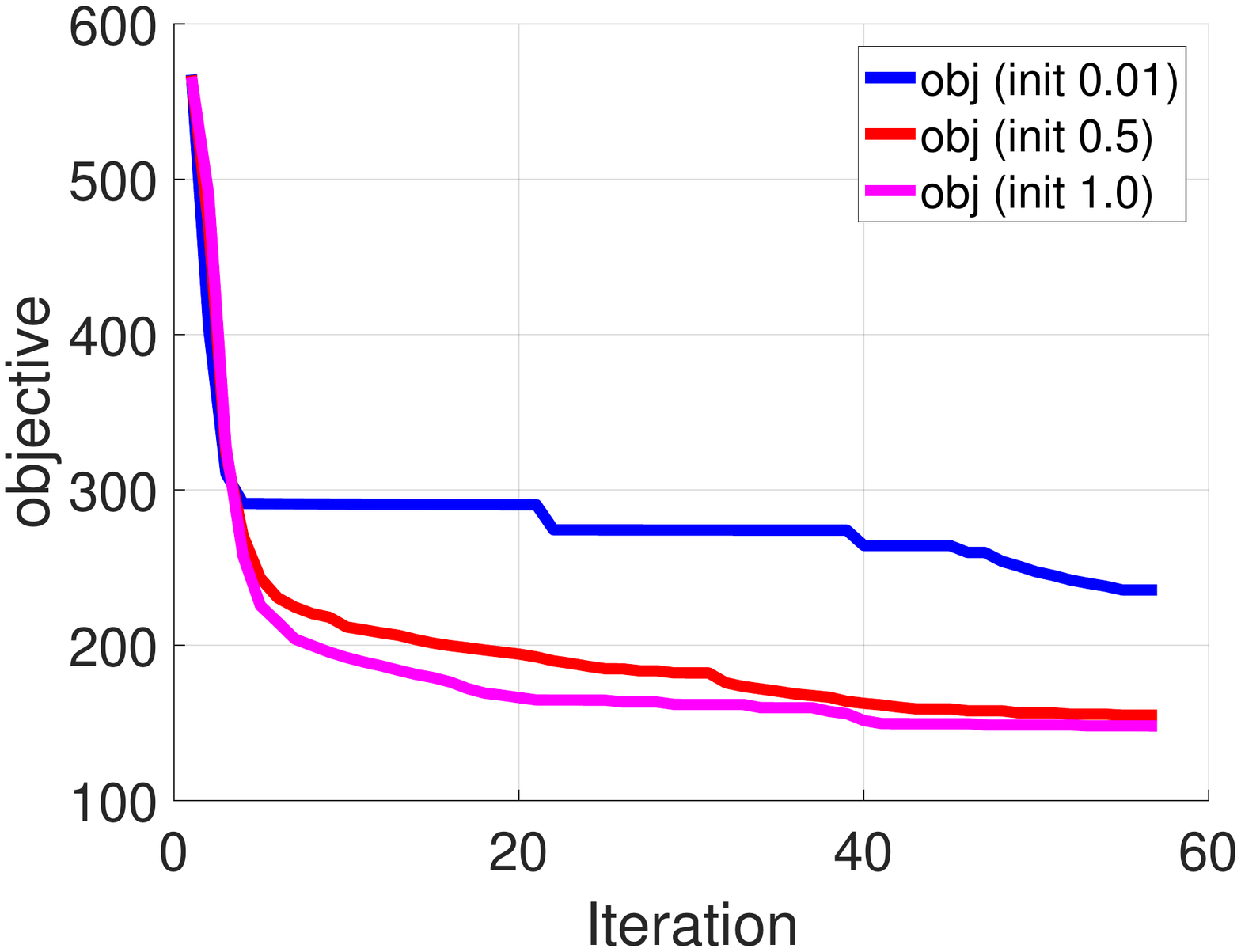}}
\subfigure[]{\includegraphics[width=6cm,trim={0 7cm 1cm 7cm},clip]{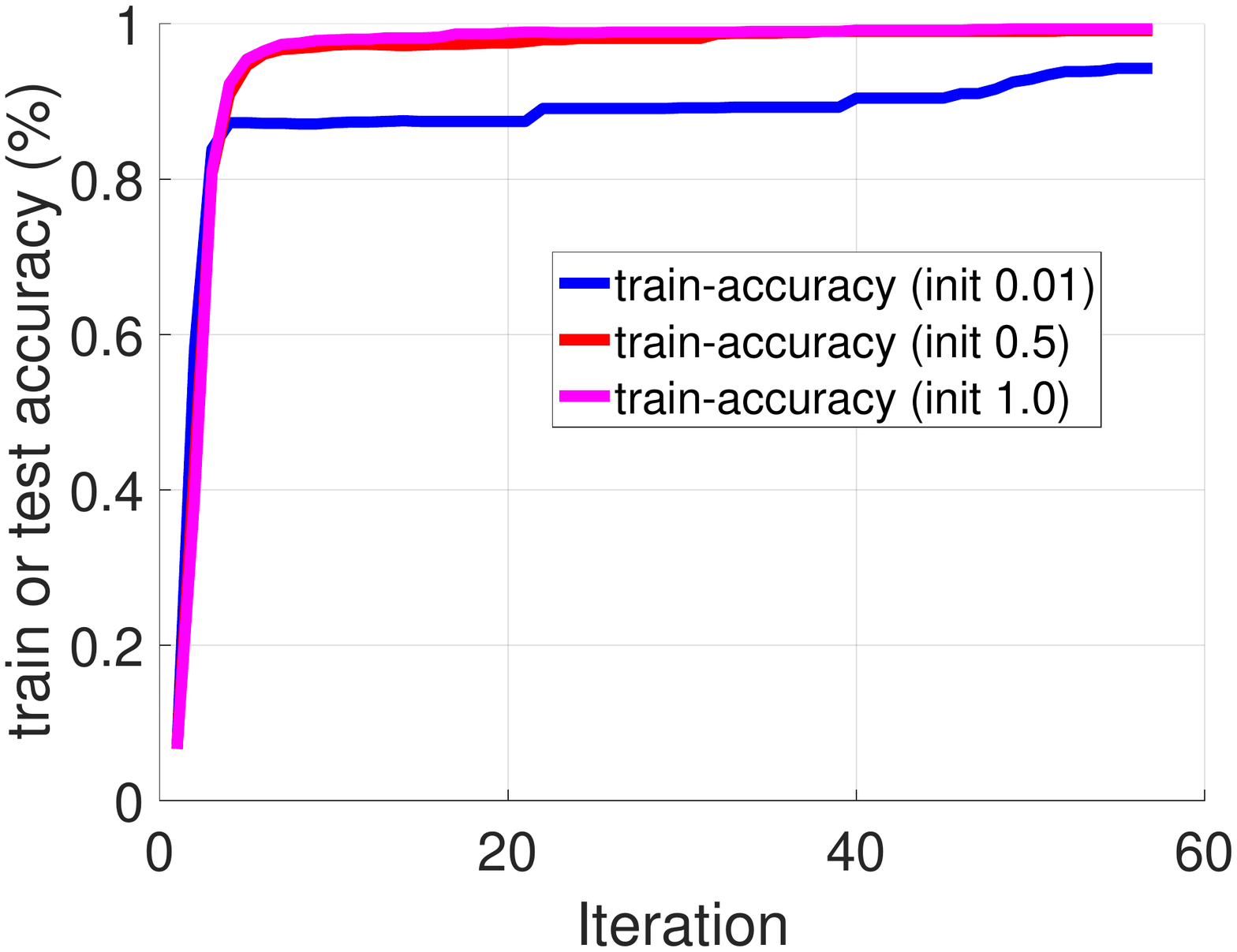}}\\
\caption{(a) Trajectory of $\alpha$ and $\beta$ over BCD iterations on the virus dataset, (b) shows the respective objective descent, and (c) shows the training set accuracy. We plot for various initializations of $\alpha$ and $\beta$.}
\label{fig:trajectory}
\end{figure*}

In this experiment, we demonstrate the BCD trajectories of $\alpha$ and $\beta$ for the IDDL-S algorithm on the Virus dataset. Specifically, in Figure~\ref{fig:trajectory}, we show how the value of $\alpha$ and $\beta$ varies as the BCD iteration progresses. In this experiment, we used 15 dictionary atoms. All experiments used the same initializations for the invariants. We also plot the corresponding objective and training accuracies. It appears that different initializations leads to disparate points of convergence. However, for all points of convergence, the objective convergence is very similar (and so is the training accuracy), suggesting that there are multiple local minima that leads to similar empirical results. We also find that initializing with $\alpha=1.0$ demonstrates slightly better convergence than other possibilities, which we observed for other datasets too. 


\begin{table*}[!htbp]
\centering
\begin{tabular}{c|c|c|c|c|c|c}
\textbf{Dataset | Classifier}   & \textbf{LE 1-NN}  & \textbf{AIRM 1-NN}    & \textbf{JBLD 1-NN} & \textbf{SVM-LE} & \textbf{IDDL}& \textbf{Variant}\\
\hline
\textbf{JHMDB}& 52.99\% & 51.87\%& 52.24\% & 54.48\% & \textbf{68.3}\% & \textbf{V}\\
\hline
\textbf{HMDB}& 29.30\% & 43.3\%& 46.3\% & 41.7\% & \textbf{55.50}\% &\textbf{N}\\
\hline
\textbf{VIRUS}& 66.67\% & 67.89\% & 68.11\% & 68.00\% & \textbf{78.39}\% & \textbf{N}\\
\hline
\textbf{BRODATZ}& 80.10\% & 80.50\% & 80.50\% & \textbf{86.80}\% & 74.10\% & \textbf{N}\\
\hline
\textbf{KTH TIPS}& 72.05\% & 72.83\%& 72.87\% & 75.59\% & \textbf{79.37}\%& \textbf{V}\\
\hline
\textbf{3D Object}&   97.4\% & 98.2\% & 95.6\% & \textbf{98.9}\% & 96.08\%& \textbf{Burg} \\
\hline
\textbf{Breast Cancer} & 87.42\%& 80.00\% & 84.00\% & 87.71\% & \textbf{90.46}\%& \textbf{Burg}\\
\hline
\textbf{Myometrium Cancer} & 80.87\% & 84.18\% & 93.20\% & 93.22\% & \textbf{94.66}\% & \textbf{Burg}\\
\end{tabular}
\caption{Comparisons against 1-NN and SVM classification. Last column shows the variant of IDDL that worked best.}
\label{tab:1stlevel}
\end{table*}

\subsection{Comparisons to Standard Measures}
In this experiment, we compare the IDDL (ridge regression) to the standard similarity measures on SPD matrices including log-Euclidean Metric~\cite{arsigny2006log}, AIRM~\cite{pennec2006}, and JBLD~\cite{cherian2013jensen}. We report 1-NN classification performance on these baselines. 
In Table~\ref{tab:1stlevel}, we report the performance of these schemes. As a rule of thumb (and also  supported empirically by cross-validation studies on our datasets), for a $C$-class problem, we chose $5C$ atoms in the dictionary. Increasing the size of the dictionary seems not helping in most cases. 
We also report a discriminative baseline by training a linear SVM on the log-Euclidean mapped SPD matrices. The results reported in Table~\ref{tab:1stlevel} demonstrates the advantage of IDDL against the baselines, where the benefits can go over more than 10\% in some cases (such as the JHMDB and virus). 

\begin{table*}[htbp]
\centering
\begin{adjustbox}{width=1.0\linewidth}
\renewcommand{\arraystretch}{1.2} 
\begin{tabular}{c|c|c|c|c|c|c|c|c|c|c}
\small{\textbf{Dataset/Method}}  & \textbf{LEML} & \textbf{$\textrm{kSP}_{LE}$}  & \textbf{$\textrm{kSP}_{JBLD}$} & \textbf{kLLC} & \textbf{RSPDL} & \textbf{IDDL-S} & \textbf{IDDL-V} & \textbf{IDDL-N} & \textbf{IDDL-A} & \textbf{IDDL-B} \\
\hline
\textbf{JHMDB} & 58.85\% & 55.97\%  &  44.40\% & 57.46\% & 57.5\% & 67.10\% & \tb{68.3\%} & 67.20\% & 61.19\% & 61.01\%\\
\hline 
\textbf{HMDB} & 52.15\% & 44.9\% & 28.43\% & 40.20\% & 21.0\% & 52.30\% & 57.3\% & \tb{58.6\%} & 43.20\% & 46.94\%\\
\hline 
\textbf{VIRUS}& 74.60\% & 68.00\% & 57.84\% & 70.91\% & 60.8\% & 76.48\% & 77.74\% & \textbf{79.44\%} & 78.33\% & 78.37\%\\
\hline
\textbf{BRODATZ}& 47.15\% & 55.00\% & 65.19\% & 70.00\% & 74.9\% & 72.50\% & 73.2\% & 77.10\% & 62.63\% & \textbf{79.44\%}\\
\hline
\textbf{KTH TIPS} & 79.25\% & 77.18\% & 69.92\% & 73.96\% & 64.5\% & 78.68\% & 79.37\% & \textbf{79.67\%} & 78.80\% & 78.36\%\\
\hline
\textbf{3D Object} & 87.56\% & 59.26\% & 72.45\% & 87.40\% & 80.0\% & 89.17\% & 94.07\% & 92.57\% & 87.90\% & \textbf{96.08\%}\\
\hline
\textbf{Breast Cancer} & 83.18\% & 76.34\% & 71.67\% & 82.32\% & 74.2\% & 89.99\% & 90.00\% & 90.02\% & 88.00\% & \textbf{90.46\%}\\
\hline
\textbf{Myometrium Cancer} & 90.94\% & 88.69\% & 86.80\% & 88.74\% & 87.0\% & 93.41\% & 93.30\% & 90.24\% & 93.99\% & \textbf{94.66\%}\\
\end{tabular}
\end{adjustbox}
\caption{Comparisons against state of the art. IDDL-A and IDDL-B refers to IDDL-AIRM and IDDL-Burg respectively. Refer to Section~\ref{sec:variants_IDDL} for details of other abbreviations.}
\label{tab:SoA}
\end{table*}
\comment{
\begin{table}[!htbp]
\centering
\begin{adjustbox}{width=1.0\linewidth}
\renewcommand{\arraystretch}{1.2} 
\begin{tabular}{c|c|c|c|c|c|c}
\textbf{Dataset | Method}  & \textbf{LEML} & \textbf{$\textrm{kSP}_{LE}$}  & \textbf{$\textrm{kSP}_{JBLD}$}    & \textbf{kLLC} & \textbf{IDDL} & \textbf{Variant}\\
\hline
\textbf{JHMDB} & 58.85\% & 55.97\%  &  44.40\% & 57.46\% & \textbf{68.3}\% &\textbf{V}\\
\hline 
\textbf{HMDB} & 52.15\% & 44.9\% & 28.43\% & 40.20\% & \textbf{55.5}\% &\textbf{N} \\
\hline 
\textbf{VIRUS}& 74.60\% & 68.00\% & 57.84\% & 70.91\% & \textbf{78.39}\% & \textbf{N}\\
\hline
\textbf{BRODATZ}& 47.15\% & 55.00\% & 65.19\% & 70.00\% & \textbf{74.10}\% &\textbf{N}\\
\hline
\textbf{KTH TIPS} & 79.25\% & 77.18\% & 69.92\% & 73.96\% & \textbf{79.37}\%& \textbf{V}\\
\hline
\textbf{3D Object} & 87.56\% & 59.26\% & 72.45\% & 87.40\% & \textbf{96.08}\% & \textbf{Burg}\\
\hline
\textbf{Breast Cancer} & 83.18\% & 76.34\% & 71.67\% & 82.32\% & \textbf{90.46}\% & \textbf{Burg}\\
\hline
\textbf{Myometrium Cancer} & 90.94\% & 88.69\% & 86.80\% & 88.74\% & \textbf{94.66}\% &\textbf{Burg}\\
\end{tabular}
\end{adjustbox}
\caption{Comparisons against state of the art. Last column shows the variant of IDDL that performed the best.}
\label{tab:SoA}
\end{table}
}

\begin{table}[]
\centering
\begin{tabular}{c|c|c|c|c}
&\multicolumn{2}{c|}{Ridge Regression} & \multicolumn{2}{c}{Structured-SVM}\\
\hline
Datasets       &  IDDL\_V & IDDV\_N & IDDL\_V & IDDL\_N \\\hline
JHMDB          & 68.3\% & 67.2\%   & \textbf{69.2}\%   & 65.8\%   \\\hline
HMDB           & 57.3\% & \textbf{58.6}\%   & 56.7\%   & 53.8\%   \\\hline
VIRUS          & 77.4\% & 79.4\%  & \textbf{79.6}\%   & 78.6\%   \\\hline
BRODATZ        & 73.2\% & 77.1\%  & 81.5\%   & \textbf{82.1}\%   \\\hline
KTH TIPS2       & \textbf{79.4}\% & 79.7\%  & 71.3\%   & 73.7\%   \\\hline
3D Object      & 94.1\% & 92.3\%  & \textbf{98.3}\%   & 98.2\%   \\\hline
Breast   & \textbf{90.0}\% & 88.0\%  & 78.1\%   & 76.7\%   \\\hline
Myometrium  & \textbf{93.3}\% & 90.2\% & 89.8\%   & 89.2\%  
\end{tabular}
\caption{Comparisons between IDDL variants for ridge regression and structured SVM alternatives.}
\label{tab:ssvm_ridge}
\end{table}

\subsection{Comparisons to the State of the Art}
We compare IDDL to the following popular methods that share similarities to our scheme, namely (i) Log-Euclidean Metric learning (LEML)~\cite{huang2015log}, (ii) kernelized Sparse Coding~\cite{harandi2014bregman} that uses log-Euclidean metric for sparse coding SPD matrices ($kSP_{LE}$), (iii) kernelized sparse coding using JBLD ($kSP_{JBLD}$), (iv) kernelized locality constrained coding~\cite{harandi2015riemannian}, and Riemannian dictionary learning and sparse coding (RSPDL)~\cite{cherian2016riemannian}. For IDDL, we chose the variant from Figure~\ref{fig:comp1st} that performed the best on the respective dataset (refer to the last column for the IDDL-variant). Our results are reported in Table~\ref{tab:SoA}. Again we observe that IDDL performs the best amongst all the competitive schemes, clearly demonstrating the advantage of learning the divergence and the dictionary. Note that comparisons are established by considering the same number of atoms for all schemes and fine-tuning the parameters of each algorithm (e.g., the bandwidth of the RBF kernel in $kSP_{JBLD}$) using a validation subset of the training set. 
As for LEML, we increased the number of pairwise constraints until the performance hit a plateau. 


In Table~\ref{tab:ssvm_ridge}, we further compare our best results on these datasets against our IDDL-V and N variants using the structured-SVM objective. We used the predicted label of the IDDL classifier (ridge regression or the structured SVM) for evaluation on all the datasets we use, except HMDB. On the HMDB dataset, we found that training an additional lib-linear SVM solver on the embeddings produced by our trained ABLD model performed better ($\sim$ 1\% better). This observation is perhaps unsurprising, given that we use 102-D covariance descriptors for this dataset (largest among our datasets), which are often (nearly) ill-conditioned, and thus may prevent the ABLD classifier from achieving the best performances in the non-convex learning setup, thereby producing embeddings that may be noisy.  From Table~\ref{tab:ssvm_ridge}, it is clear that our SSVM formulation is better on some datasets, for example, on Brodatz, and 3D object datasets, there is a substantial gain of nearly 5\%, while on others the performance is similar to ridge regression objective. On the smaller datasets, such as Breast and Myometrium cancer, ridge regression is much better.

\subsection{Evaluation of Clustering Objective}
Now, we present experiments on the aforementioned benchmarks using our $\alpha\beta$-KMeans clustering framework. To evaluate the quality of clustering, we use the standard F1-Score. Before presenting comparisons to other popular clustering schemes on SPD matrices, we study the empirical properties of our formulation next. 

%
We compare the quality of clustering against (i) dimensionality of the input matrices, and (ii) number of true clusters, and iii) time taken for the relevant updates. For this experiment, we use synthetic datasets generated using the code from~\cite{cherian2016bayesian}, which produces Wishart SPD matrix clusters for $k$ arbitrarily parameterized Wishart distributions; $k$ being the number of true data clusters.

\subsubsection{Increasing Matrix Dimensionality}
\begin{figure*}[htbp]
\begin{center}
\subfigure[]{\label{fig:inc_dim_acc}\includegraphics[width=4.4cm]{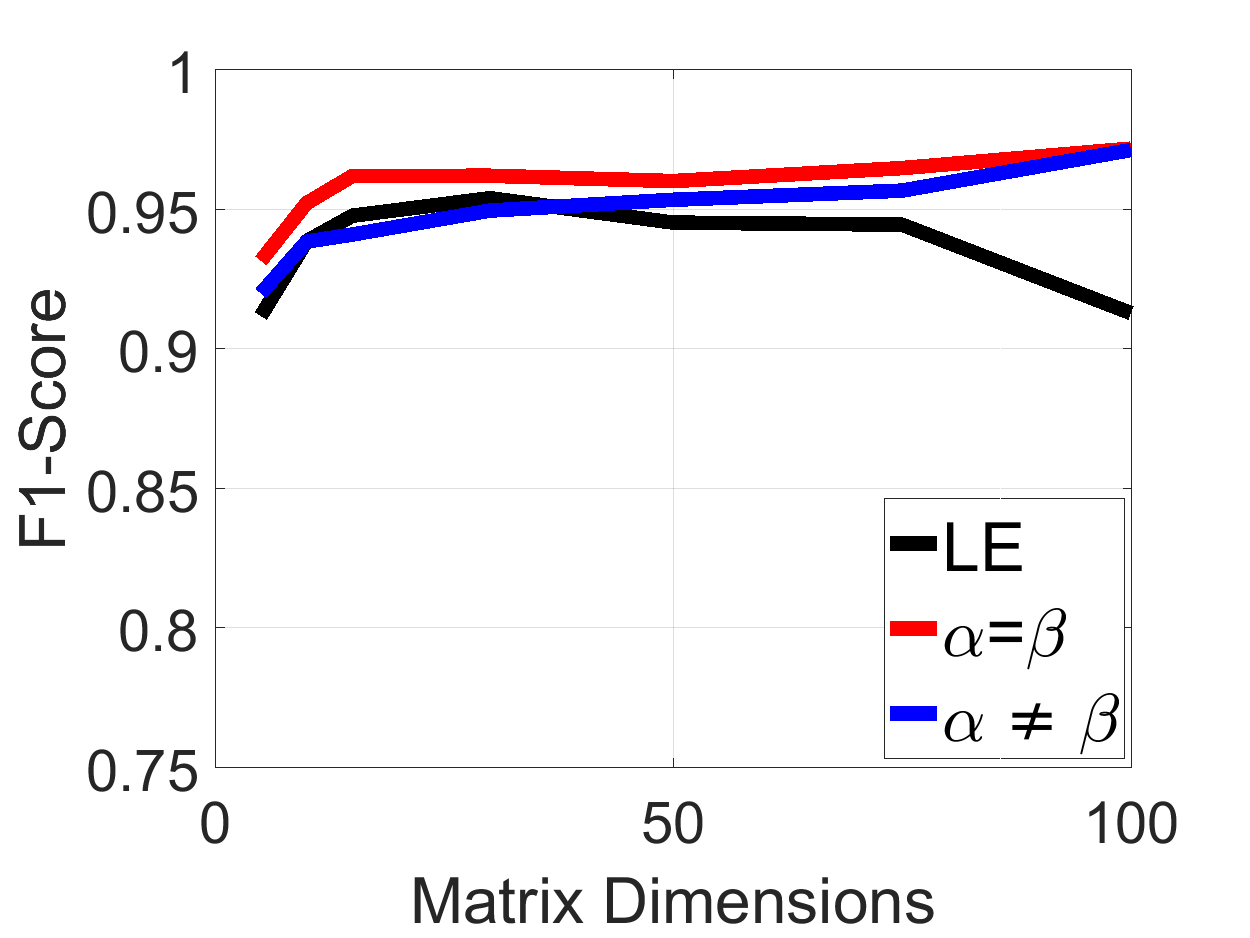}}
\subfigure[]{\label{fig:inc_dim_time}\includegraphics[width=4.4cm]{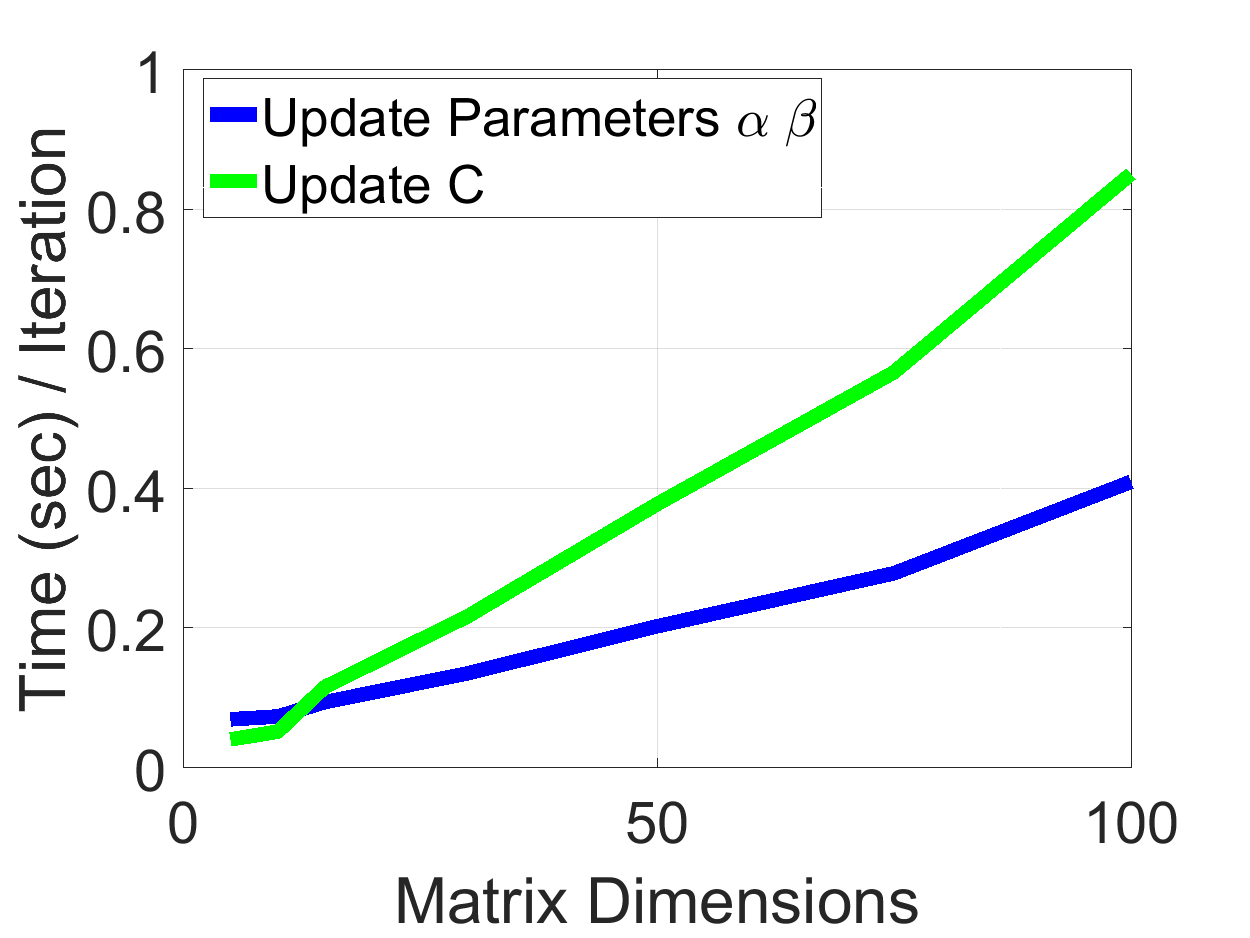}}
\subfigure[]{\label{fig:inc_clust_acc}\includegraphics[width=4.4cm]{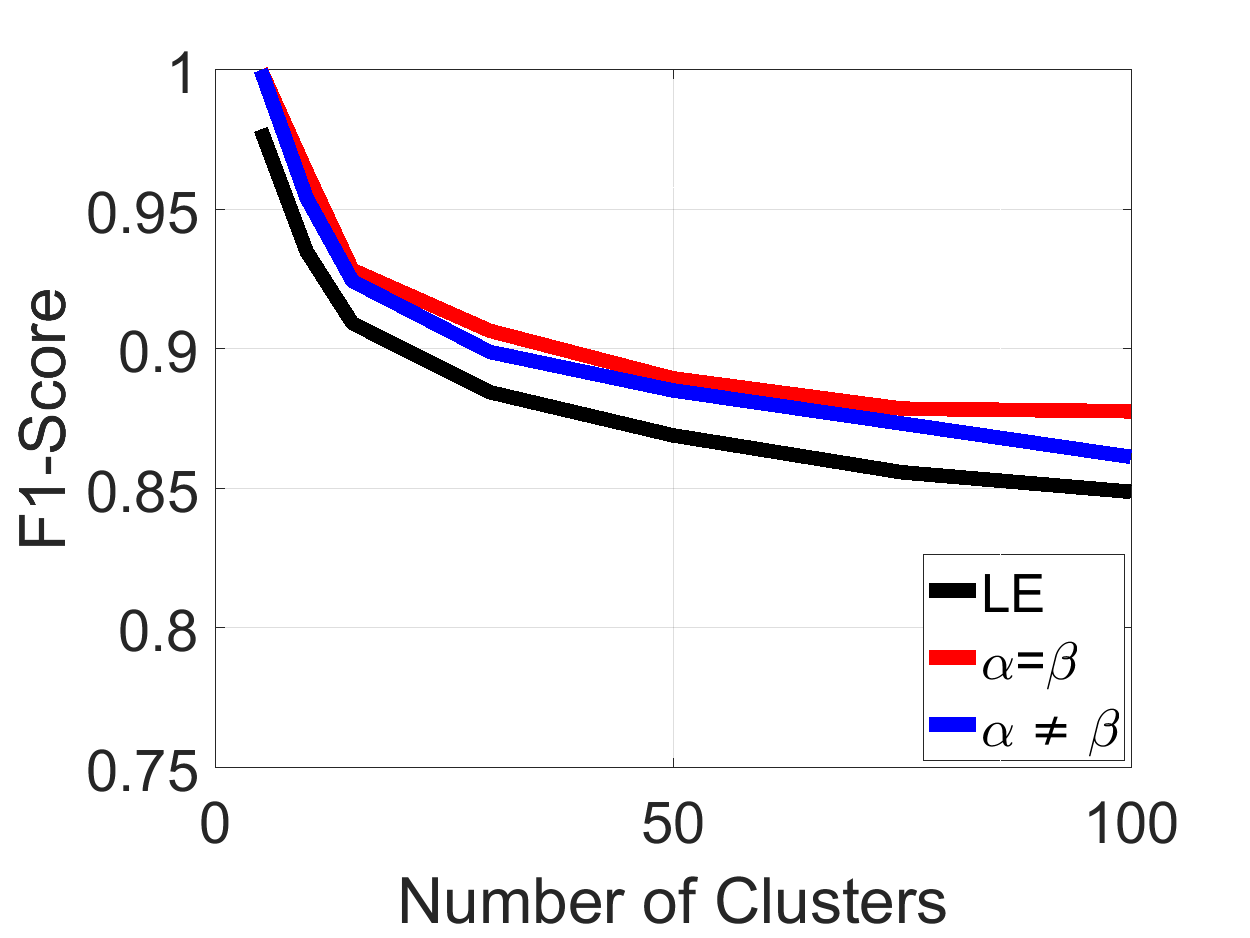}}
\subfigure[]{\label{fig:inc_clust_time}\includegraphics[width=4.4cm]{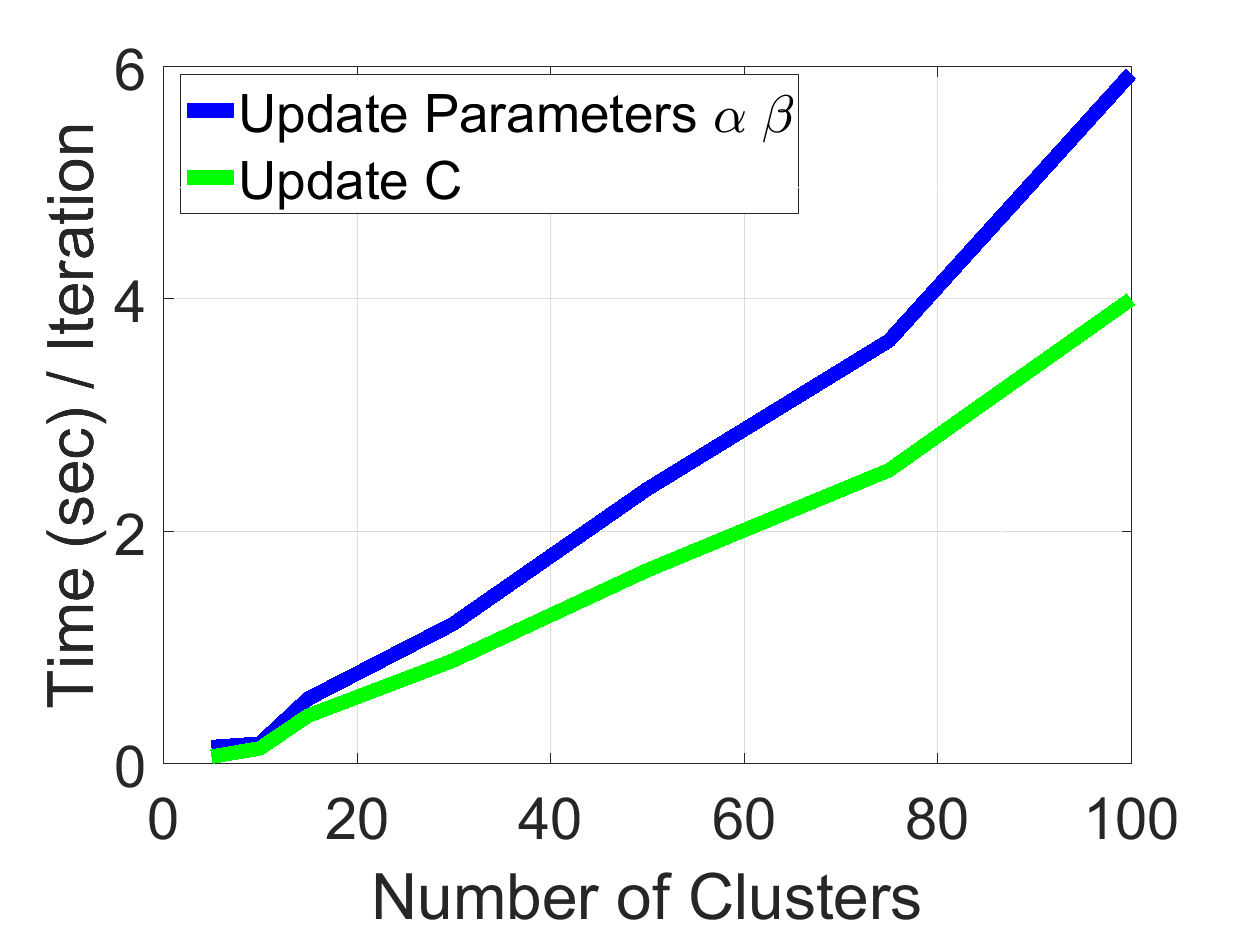}}
\caption{Sensitivity of $\alpha\beta$-KMeans against ~\ref{fig:inc_dim_acc} increasing dimensionality in the range $[5, 100]$. The blue and red lines correspond to $\alpha\beta$-KMeans with $\alpha=\beta$ and $\alpha\neq\beta$ respectively, while the black line corresponds to LE-KMeans.~\ref{fig:inc_dim_time} Time required for each iteration of updating parameters $\alpha\beta$ (blue line) and centroids (green line).~\ref{fig:inc_clust_acc} and~\ref{fig:inc_clust_time} show the same for increasing number of clusters.}
\label{fig:sens_dim}
\end{center}
\end{figure*}

For this experiment, we generate synthetic SPD datasets of dimensionality $d$, where $d \in \{5, 15, 30, 50, 75, 100\}$ corresponding to $k=15$ clusters and using fifty samples per class. Figure~\ref{fig:inc_dim_acc} summarizes the computed F1-scores averaged across ten runs. We can clearly see that the accuracy of $\alpha\beta$-KMeans is not impacted much by the increasing dimensions of the input matrices, while both variants consistently outperform the baseline of LE-KMeans. Figure~\ref{fig:inc_dim_time} present the time taken for a single iteration of each optimization component of $\alpha\beta$-KMeans, which looks approximately linear.

\subsubsection{Increasing Number of Clusters}
Next, we test the robustness of the $\alpha\beta$-KMeans with increasing number of true data clusters $k$, for $k \in \{2, 5, 10, 20, 50, 100\}$. For this experiment, we keep the dimension of the SPD matrices fixed to $d=10$ and use twenty five samples per true data class. Figure~\ref{fig:inc_clust_acc} summarizes the F1-Score of $\alpha\beta$-KMeans averaged across ten runs for an increasing number of clusters. We can infer that both variants are negatively affected by large increases in the number of clusters, nevertheless, the our performance is consistently higher than that of the LE-KMeans baseline. In addition, as depicted in Figure~\ref{fig:inc_clust_time}, there is an increasing overall trend in the time required for all components of $\alpha\beta$-KMeans (compared to Figure~\ref{fig:inc_dim_time}, due to the time required to iterate through the different clusters.

\subsubsection{Empirical Convergence Analysis\label{sec:EmpAnalysis}}
\begin{figure}[!htbp]
\begin{center}
\includegraphics[width=0.75\linewidth]{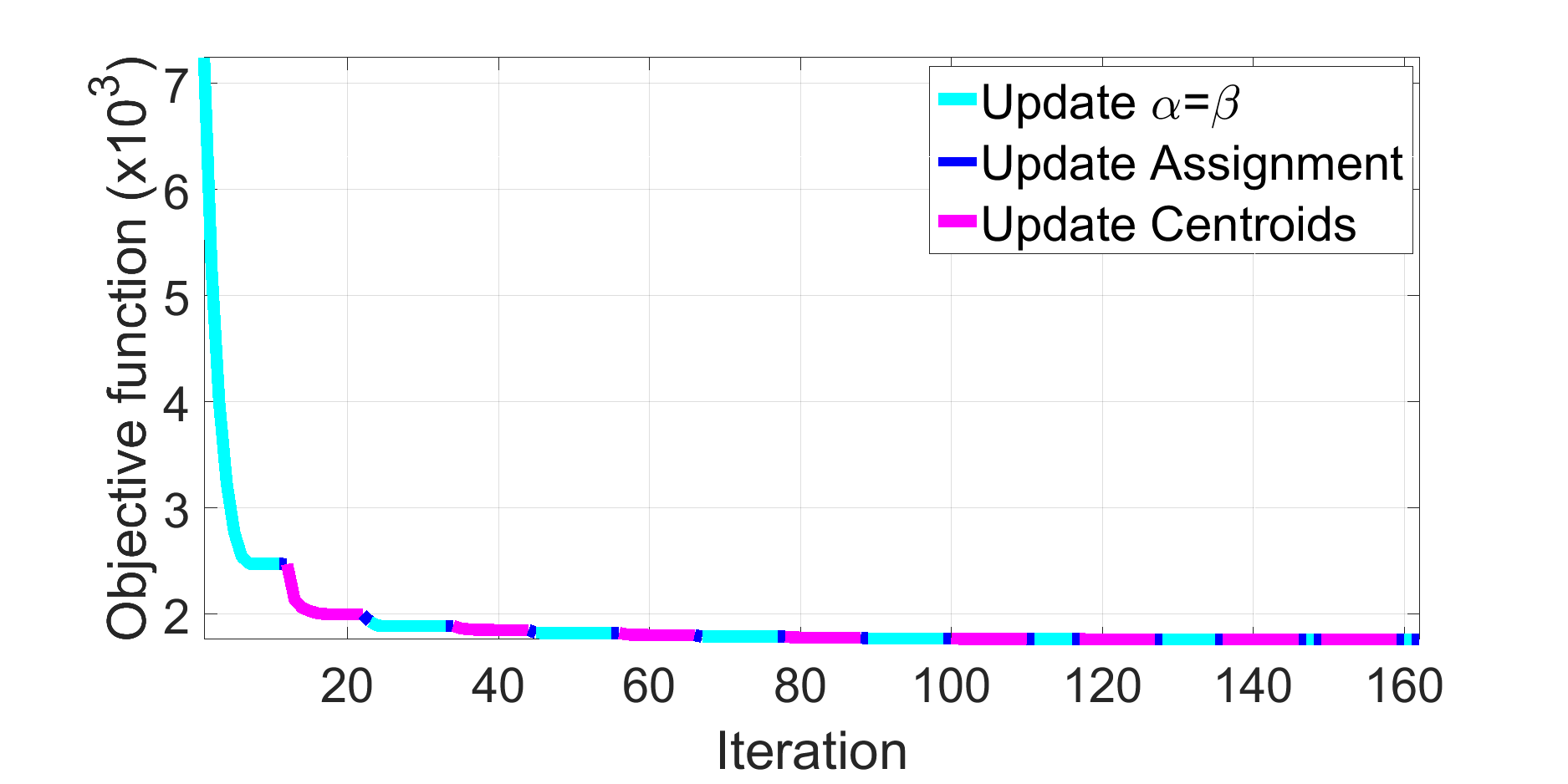}
\caption{Convergence plot of the objective function~\ref{eq:abkmeans} for the myometrium cancer dataset and $\alpha=\beta$. Cyan line segments correspond to iterations of updating the divergence parameters, blue segments correspond to updating the clustering assignment and magenta segments correspond to iterations of updating the centroids.}
\label{fig:empconv}
\end{center}
\end{figure}

Now, we empirically study the convergence of $\alpha\beta$-KMeans. We select to present this analysis on the Myometrium cancer dataset nevertheless, the results remain consistent among the different datasets. Figure~\ref{fig:empconv} illustrates the convergence of the BCD scheme discussed in Section~\ref{sec:proposed_method} for $\alpha\beta$-KMeans with $\alpha=\beta$. Even though the objective is non-convex, it is apparent that the  convergence is satisfactory. We run the scheme until more than 99.9\% of the clustering assignments remain unchanged between two successive clustering steps. 

\subsubsection{Comparisons to Variants of KMeans}
Comparisons are first established against two popular variants of KMeans for SPD matrices; LE-KMeans and Karcher Means. Table~\ref{tab:SoA} summarizes the experiments evaluating the performance of the $\alpha\beta$-KMeans in a pure clustering setup. The first and second columns correspond to the F1-Score achieved by LE-KMeans and Karcher Means respectively. The two proposed variants of $\alpha\beta$-KMeans are depicted in columns three ($\alpha=\beta$) and four ($\alpha\neq\beta$). For each dataset, we average our results across ten different runs to alleviate the effect of initializations. We can clearly see that the two variants of $\alpha\beta$-KMeans consistently outperform the competing schemes underlying the merits of learning the measure while clustering the data.

\begin{table}[!htbp]
\centering
\begin{adjustbox}{width=1\linewidth}
\renewcommand{\arraystretch}{1.2} 
\begin{tabular}{c|c|c|c|c}
\textbf{Dataset | Method}  & \textbf{LE} & \textbf{Karcher} & $\valpha\vbeta$-\textbf{E}& $\valpha\vbeta$-\textbf{NE}\\
\hline
\textbf{VIRUS}& 0.248 & 0.254 & 0.252 & \textbf{0.257}\\
\hline
\textbf{BRODATZ}& 0.353 & 0.366 & 0.378 & \textbf{0.381} \\
\hline
\textbf{KTH TIPS} & 0.379 & 0.400 & \textbf{0.429} & 0.419 \\
\hline
\textbf{Prostate Cancer} & 0.578 & 0.594 & \textbf{0.679 }& 0.660 \\
\hline
\textbf{Myometrium Cancer} & 0.737 & 0.661 & 0.778 & \textbf{0.779} \\
\end{tabular}
\end{adjustbox}
\caption{F1-Score based comparisons against different KMeans variants.}
\label{tab:SoA}
\end{table}

\section{Discussions}
Our experiments show that learning the divergence and the parameters of the respective task demonstrate superior performances on all the datasets we used. That said, there are also some challenges one may need to circumvent when using the setup. The main limitation of our approach is the non-convexity of our objective; that precludes a formal analysis of the convergence. A further limitation is that the gradient expressions involve matrix inversions and may need careful regularizations to avoid numerical instability. We also note that the AB divergence has a discontinuity at the origin, which needs to be accounted for when learning the parameters. Further, from our experimental analysis, it looks like there is no single variant of IDDL (amongst IDDL-S, IDDL-V, IDDL-N, IDDL-A, and IDDL-B) that consistently performs the best for all datasets. However, with the possibility of learning alpha-beta, we would think the most generalized variant IDDL-N with the structurd-SVM formulation might perhaps be the best choice for any application as it can plausibly learn all the alternatives.

As for our clustering setup, we found that it is essential to use a regularizer on $\alpha$ and $\beta$; in the absence of which, the optimization was seen to diverge, the parameters taking very large values leading to irrecoverable numerical deficiencies. As noted earlier, we found quadratic regularizers on $\alpha,\beta$ yielded good results. Exploring other forms, such as polynomials on $\alpha$ and $\beta$, or robust priors such as the Huber loss, is left as future work. From our experiments on real data, we found beneficial small additive perturbations on the diagonal of the SPD matrices (to make them strongly positive definite). On all our datasets, we found each block of updates using RCG converged in about 5-10 steps. Surprisingly, the proposed BCD scheme is seen to converge much faster for the $\alpha\neq \beta$-case in comparison to $\alpha=\beta$, when centroids are initialized using the LE-KMeans rather than randomly selecting samples from each dataset. This faster convergence is perhaps because of the more degrees of parameter freedom and the conditioning of the matrices. 

\section{Conclusions}
In this paper, we proposed a novel framework unifying the problem of information divergence learning and standard machine learning tasks, such as dictionary learning, ridge regression, classification, and clustering
on SPD matrices. We leveraged the recent advances in information geometry for this purpose, namely using the $\alpha\beta$-logdet divergence. We formulated objectives for jointly learning the divergence and the respective task specific objectives, and showed that it can be solved efficiently using optimization methods on Riemannian manifolds in an end-to-end manner. Experiments on eight computer vision datasets demonstrate superior performance of our approach against alternatives.

\section{Acknowledgements}
This material is based upon work supported
by the National Science Foundation through grants   \#SMA-1028076, \#CNS-1338042,
\#CNS-1439728,   \#CNS-1514626, \#CNS-1939033, and \#CNS-1919631. Anoop Cherian was
funded by the Australian Research Council Centre of Excellence
for Robotic Vision (\#CE140100016).  Research reported in this publication was supported by the National Cancer Institute of the NIH under Award Number R01CA225435.

\appendix
\noindent\textbf{Proof of Theorem~\ref{theor:logdet_der}.}
To simplify the notation, let $S=\mA^{\half}$. Note that the eigenvalues (and hence the $\logdet$) of $\mA\mB$ is the same as that of $\mA^{\half}\mB\mA^{\half}$, however, the latter is symmetric and thus keeps $B$ symmetric, when using it in a gradient descent scheme. Thus, we will use this form. Then,
\begin{equation}
\label{eq:18}
\logdet\left[p\left(\mS\mB\mS\right)^{q}+\eye{d}\right] = \trace{\Log \left(p\left(\mS\mB\mS\right)^{q}+\eye{d}\right)}.
\end{equation}

\noindent Using Taylor series expansion of $\Log$:
\begin{align}
\eqref{eq:18}&=\trace{p (\mS\mB\mS)^{q}-\frac{p^2(\mS\mB\mS)^{2q}}{2}-\cdots}.\label{eq:19}\\
\grad{\mB}~\eqref{eq:19}&=pq\mS(\mS\mB\mS)^{q-1}S - q p^2\mS(\mS\mB\mS)^{2q-1} \mS +\cdots \notag \\
& = pq \mS\inv{(\mS\mB\mS)}\left(\mS\mB\mS\right)^{q}\left(\eye{d} - p\left(\mS\mB\mS\right)^{q}+\cdots\right)\mS \notag
\end{align}
Recall that, the middle term is the MacLaurin series expansion:
\begin{equation}
   \left(\eye{d} - p\left(\mS\mB\mS\right)^{q}+\cdots\right)  = \inv{(\eye{d} + p \left(\mS\mB\mS\right)^{q})}\notag
\end{equation}
substituting for which we get our desired result. Note that, the series expansions we use in the proof are valid only when $p\enorm{SBS}_2\leq 1$, which can be achieved via rescaling or normalizing our data.
$\blacksquare$



\ifCLASSOPTIONcaptionsoff
  \newpage
\fi

{\small
\bibliographystyle{IEEEtran}
\bibliography{ABLD-PAMI.bib}
}







\begin{IEEEbiography}[{\includegraphics[width=2.8cm,height=2.8cm,clip]{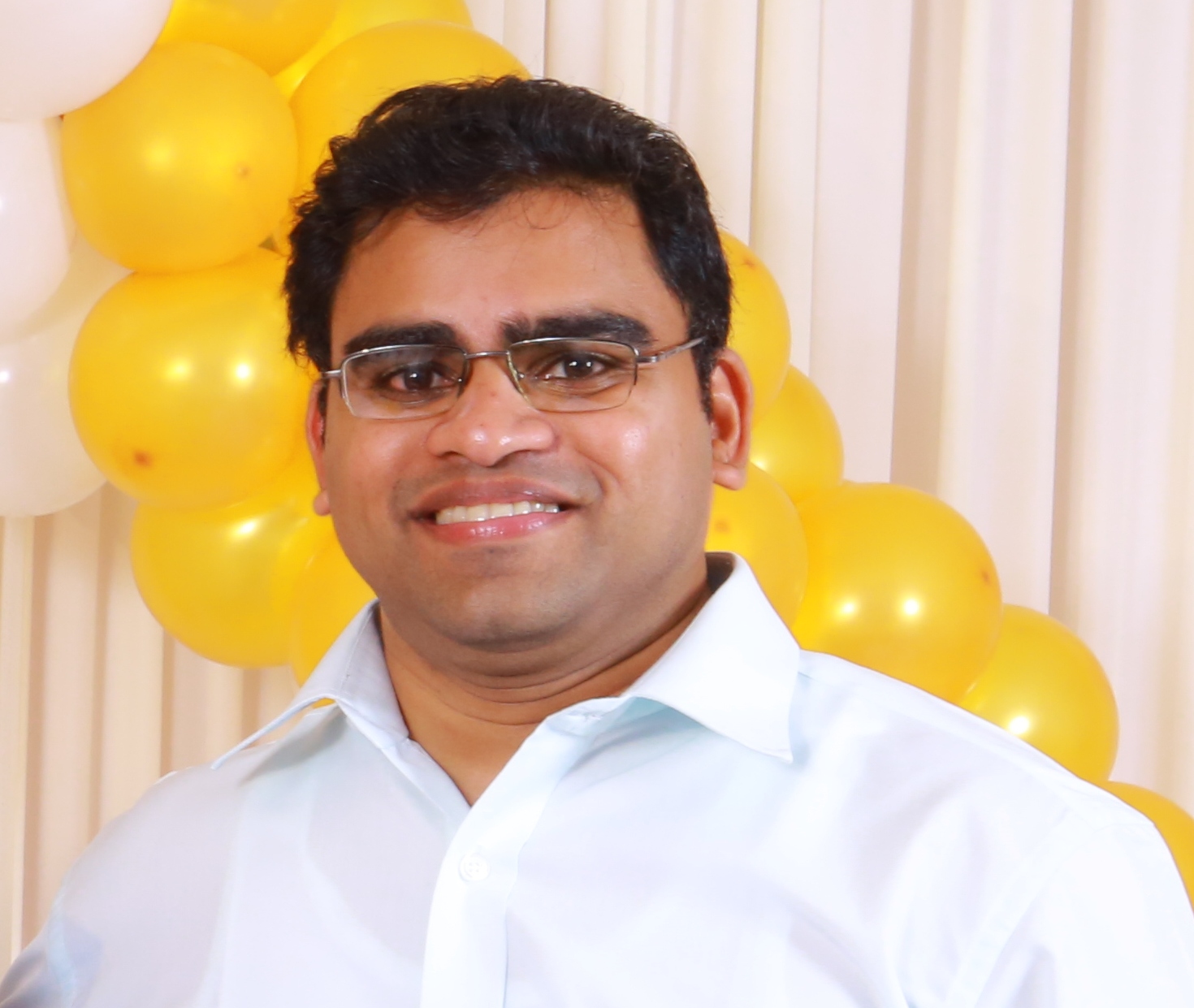}}]{Anoop Cherian}
is a Principal Research Scientist with Mitsubishi Electric Research Labs (MERL), Cambridge, MA and an Adjunct Researcher with the Australian National University. He received his M.S. and Ph.D. degrees in computer science from the University of Minnesota, Minneapolis in 2010 and 2013, respectively. He was a postdoctoral researcher in the LEAR group at Inria, Grenoble from 2012-2015, and a Research Fellow at the Australian National University from 2015-2017. Anoop has broad interests in machine learning, deep learning, and computer vision, and has authored more than 50 scientific articles. He is also the recipient of several awards, including the Best Student Paper Award at ICIP, 2012.
\end{IEEEbiography}
\begin{IEEEbiography}
[{\includegraphics[width=1in,height=1.15in,clip,keepaspectratio]{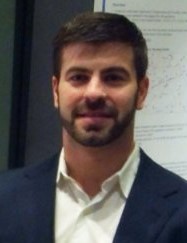}}]{Panagiotis Stanitsas} received his Diploma of Engineering in Civil Engineering from the University of Patras, Greece in 2010. He received his M.S. in Project Management and Transportation from the University of Patras, Greece in 2011 and his M.Sc in Civil Engineering from the University of Minnesota in 2013. He received his Ph.D. from the department of Computer Science and Engineering at the University of Minnesota in 2017. His research interests include machine learning and computer vision with emphasis on active schemes for learning. He was awarded with an award for excellence in Transportation from the University of Patras in 2010, while he has also received the interdisciplinary Matthew J. Huber award of excellence at the University of  Minnesota in 2013.
\end{IEEEbiography}
\begin{IEEEbiography}[{\includegraphics[width=1in,height=1.25in,clip,keepaspectratio]{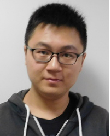}}]{Jue Wang}
is a PhD student with the Research School of Engineering at the Australian National University since 2016. He is also associated with CSIRO's Data61 in Australia. From 2010-2014, he received his double bachelor degree (honors) in Electronic Engineering from Australian National University and Beijing Institute of Technology. His research interest are in the area of computer vision and machine learning.
\end{IEEEbiography}
\begin{IEEEbiography}[{\includegraphics[width=2.8cm,clip,keepaspectratio]{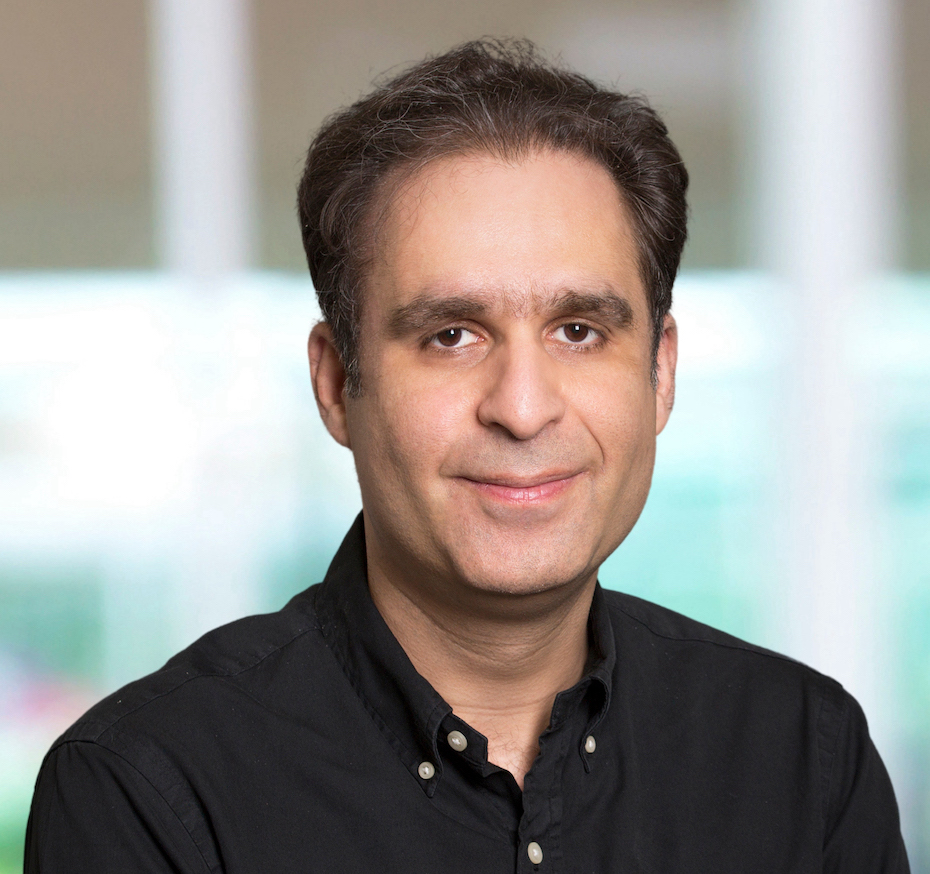}}]{Mehrtash Harandi}
is a Senior Lecturer with the Department of Electrical and Computer Systems Engineering at Monash University. He is also a contributing research scientist in the Machine Learning Research Group (MLRG) at Data61/CSIRO and an associated investigator at the Australian Center for Robotic Vision (ACRV). His current research interests include theoretical and computational methods in machine learning, computer vision, signal processing, and Riemannian geometry.
\end{IEEEbiography}


\begin{IEEEbiography}
[{\includegraphics[width=1in,height=1.15in,clip,keepaspectratio]{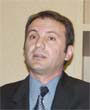}}]{Vassilios Morellas}
received his diploma degree in Mechanical Engineering from the National Technical University of Athens, Greece, his MSME degree from Columbia University, NY and his Ph.D. degree from the department of Mechanical Engineering at the University of Minnesota. He is Research Professor in the department of Electrical  and Computer Engineering and Executive Director of the NSF Center for Robots and Sensors for the Human Well-Being. His research interests are in the area of geometric image processing, machine learning, robotics and sensor integration.
\end{IEEEbiography}


\begin{IEEEbiography}
[{\includegraphics[width=1in,height=1.15in,clip,keepaspectratio]{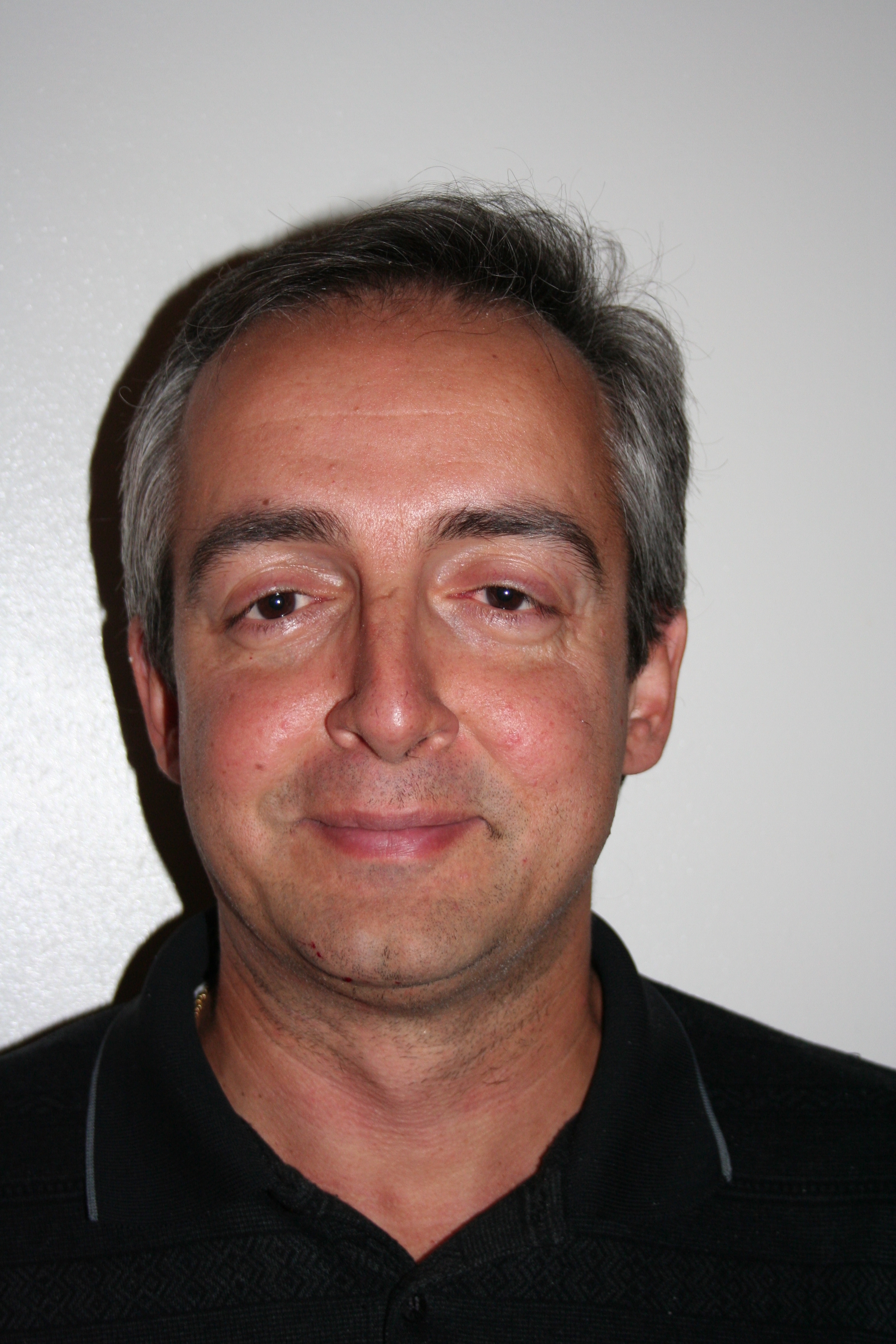}}]{Nikos Papanikolopoulos}
(IEEE Fellow) received his Diploma of Engineering in Electrical and Computer Engineering, from the National Technical University of Athens in 1987. He received his M.S. in 1988 and Ph.D. in 1992 in Electrical and Computer Engineering from Carnegie Mellon University. His research interests include computer vision, robotics, sensors for transportation and precision agriculture applications,  and control systems.  He is the Director of the Minnesota Robotics Institute and the McKnight Presidential Endowed Professor at the University of Minnesota. He has also received numerous awards including the 2016 IEEE RAS George Saridis Leadership Award in Robotics and Automation.
\end{IEEEbiography}

\end{document}